\documentclass[10pt,journal,cspaper,final,twocolumn,compsoc]{./IEEEtran}
\pdfoutput=1
\usepackage[pdftex]{graphicx}
\usepackage[cmex10]{amsmath} 

\usepackage{float}
\usepackage{bm}
\usepackage{adjustbox}     
 
\usepackage{enumitem}
\usepackage{colortbl}

\newif\ifrevfinal
\revfinalfalse
\def\rev[#1][#2]{\ifrevfinal #2 \else {\color{blue} \sout{#1}} {\bf \color{red} #2} \fi}

\usepackage{color}
\usepackage[normalem]{ulem}

\def\etal{et al.\ }

\def\<{\langle}
\def\>{\rangle}

\def\be{\begin{equation}}
\def\ee{\end{equation}}
\def\bea{\begin{eqnarray}}
\def\eea{\end{eqnarray}}
\def\fig#1{Figure~\ref{fig:#1}}
\def\tab#1{Table~\ref{tab:#1}}
\def\sect#1{Section~\ref{sec:#1}}

\def\algo#1{Algorithm~\ref{algo:#1}}

\makeatletter
\gdef\SetFigFont#1#2#3{%
  \reset@font\fontsize{10}{12pt}%
  \selectfont%
}
\let\eqnarray@=\eqnarray \let\endeqnarray@=\endeqnarray
\def\eqnarray{\bgroup\arraycolsep=2pt\eqnarray@}
\def\endeqnarray{\endeqnarray@\egroup}
\makeatother

\def\em{\slshape}

\usepackage{times,amsmath,amssymb,amsbsy,xspace}
\makeatletter
\DeclareRobustCommand\onedot{\futurelet\@let@token\@onedot}
\def\@onedot{\ifx\@let@token.\else.\null\fi\xspace}

\def\eg{\emph{e.g}\onedot} 
\def\ie{\emph{i.e}\onedot} 
\def\cf{\emph{c.f}\onedot} 
 \def\vs{\emph{vs}\onedot} 
\def\wrt{w.r.t\onedot} 
\def\etal{\emph{et al}\onedot}

\newcounter{iictr}
\def\@iia[#1]{\setcounter{iictr}{#1}\@iib}
\def\@iib{\iii[\roman{iictr}]\xspace}
\def\ii{\@ifnextchar[{\@iia}{\stepcounter{iictr}\@iib}}
\def\iii[#1]{({\em#1\/})}
\makeatother

\graphicspath{
{images/}
}

\makeatletter

\providecommand{\tabularnewline}{\\} 
\floatstyle{ruled}
\newfloat{algorithm}{tbp}{loa}
\floatname{algorithm}{Algorithm}

\ifCLASSOPTIONcompsoc
   \usepackage[nocompress]{cite} 
\else
  \usepackage{cite}
\fi
\usepackage{algorithmic} 
\usepackage{array} 

\ifCLASSOPTIONcompsoc
  \usepackage[caption=false,font=footnotesize,labelfont=sf,textfont=sf]{subfig} 
\else
  \usepackage[caption=false,font=footnotesize]{subfig}
\fi

\usepackage{url} 

 \usepackage[pagebackref=false,breaklinks=true,colorlinks,bookmarks=false]{hyperref} 

\makeatother

\def\VOC{PASCAL VOC }

\def\rev[#1][#2]{{\color{blue} \sout{#1}} {\bf \color{red} #2}}

\begin{document}

\title{Weakly Supervised Object Localization with Multi-fold Multiple Instance Learning}

\author{Ramazan~Gokberk~Cinbis, %
        Jakob~Verbeek, %
        and~Cordelia~Schmid,~\IEEEmembership{Fellow,~IEEE}%
\IEEEcompsocitemizethanks{\IEEEcompsocthanksitem R. G. Cinbis is with the Department of Computer Engineering, Bilkent University,
Ankara, Turkey. Most of this work was done when he was with LEAR team, Inria Grenoble Rh\^one-Alpes, Laboratoire Jean Kuntzmann, CNRS, Univ.\ Grenoble Alpes, France. E-mail: gcinbis@cs.bilkent.edu.tr\protect\\
\IEEEcompsocthanksitem J. Verbeek and C. Schmid are with LEAR team, Inria Grenoble Rh\^one-Alpes, Laboratoire Jean Kuntzmann, CNRS, Univ.\ Grenoble Alpes, France. E-mail: firstname.lastname@inria.fr}
\thanks{Copyright (c) 2016 IEEE. Personal use of this material is permitted. However, permission to use this material for any other purposes must be obtained from the IEEE by sending a request to pubs-permissions@ieee.org.}}

\markboth{TO APPEAR IN IEEE TRANSACTIONS ON PATTERN ANALYSIS AND MACHINE INTELLIGENCE, 2016}
{}

\IEEEtitleabstractindextext{%
\begin{abstract} 
Object category localization is a challenging problem in
computer vision. Standard supervised training requires
bounding box annotations of object instances. This
time-consuming annotation process is sidestepped in weakly
supervised learning. In this case, the supervised
information is restricted to binary labels that indicate
the absence/presence of object instances in the image,
without their locations. We follow a multiple-instance
learning approach that iteratively trains the detector
and infers the object locations in the positive training
images. Our main contribution is a multi-fold multiple
instance learning procedure, which prevents training from
prematurely locking onto erroneous object locations. This
procedure is particularly important when using high-dimensional
representations, such as Fisher vectors and convolutional
neural network features. We also propose a 
window refinement method, which improves the localization
accuracy by incorporating an objectness prior.
We present a detailed experimental evaluation using the
\VOC 2007 dataset, which verifies the effectiveness of our
approach.
\end{abstract}

\begin{IEEEkeywords}
Weakly supervised learning, object detection.
\end{IEEEkeywords}
}

\maketitle
\IEEEdisplaynontitleabstractindextext

\IEEEraisesectionheading{\section{Introduction}\label{sec:intro}}

\IEEEPARstart{O}{ver}
the last decade significant progress has been made in
object category localization, as witnessed by the \VOC
challenges~\cite{everingham10ijcv}. Training
state-of-the-art object detectors, however, requires bounding box
annotations of object instances, which are costly to acquire. 

Weakly supervised learning (WSL) refers to methods that
rely on training data with incomplete ground-truth
information to learn recognition models. For object detection, WSL from image-wide labels that indicate the presence of instances of a category in images
has recently been intensively studied as a way to remove
the need for bounding box annotations, see \eg
\cite{bagon10cvpr,chum07cvpr,crandall06eccv,deselaers12ijcv,pandey11iccv,prest12cvpr,russakovsky12eccv,shi13iccv,siva12eccv,siva11iccv,song14icml,song14nips,bilen14bmvc,wang14eccv}.
Such methods can potentially leverage the large amount of
tagged images on the internet as a data source to train
object detectors. We give an overview of the most relevant related work in \sect{related}.

Other examples of WSL include learning face recognition
models from image captions~\cite{berg04cvpr}, or subtitle
and script information~\cite{everingham09ivc}. Yet another 
example is learning semantic segmentation models from
image-wide category labels~\cite{verbeek07cvpr}.  Most WSL
approaches are based on latent variable models to account
for the missing ground-truth information. Multiple
instance learning (MIL)~\cite{dietterich97ai} handles
cases where the weak supervision indicates that at least
one positive instance is present in a set of examples.
More advanced inference and learning methods are used in
cases where the latent variable structure is more complex,
see \eg~\cite{deselaers12ijcv,shi13iccv,verbeek07cvpr}.
Besides weakly supervised training, mixed fully and weakly
supervised~\cite{blaschko10nips}, active~\cite{vijayanarasimhan11cvpr}, and semi-supervised~\cite{shi13iccv} learning and unsupervised object discovery~\cite{cho15cvpr} methods have also been explored
to reduce the amount of labeled training data for object
detector training.  In active learning bounding box
annotations are used, but requested only for images where
the annotation is expected to be most effective.
Semi-supervised learning, on the other hand, leverages
unlabeled images by automatically detecting objects in
them, and use those to better model the object appearance
variations. 

In this paper we consider WSL  to learn
object detectors from image-wide labels. We follow an MIL
approach that interleaves training of the detector with
re-localization of object instances on the positive
training images. Following recent state-of-the-art work
in fully supervised detection
\cite{cinbis13iccv}\cite{girshick14cvpr}\cite{sande14cvpr}, we
represent (tentative) detection windows using Fisher
vectors (FVs)~\cite{sanchez13ijcv} and convolutional
neural network (CNN) features~\cite{krizhevsky12nips}. As
we explain in \sect{method}, when used in an MIL
framework, the high-dimensionality of the window features
makes MIL quickly convergence to poor local optima after
initialization.  Our main contribution is a multi-fold
training procedure for MIL, which avoids this rapid
convergence to poor local optima.  A second novelty of our
approach  is the use of a ``contrastive'' background
descriptor that is defined as the difference of a
descriptor of the object window and a descriptor of the
remaining image area. The score for this descriptor of a
linear classifier can be interpreted as the difference of
scores for the foreground and background. In this manner
we direct the detector to learn the difference between
foreground and background appearances. Finally, inspired from 
the objectness prior in \cite{deselaers12ijcv}, we propose
a window refinement method that improves the
weakly supervised localization accuracy by incorporating a
category-independent objectness measure.

 We present a detailed evaluation using the VOC 2007 dataset in \sect{experiments}.
The experimental results show that our multi-fold MIL training
improves performance for both FV and CNN
features. We also show that WSL performance can be further
improved by combining the two descriptor types and
applying our window refinement method. The evaluation shows that our system obtains state-of-the-art results on VOC 2007. We also present results for VOC 2010 which was not yet used in previous work.

Part of the material presented here appeared in \cite{cinbis14cvpr}. Besides a more detailed presentation and discussion of the most recent related work,  the current paper extends it in several ways. We enhanced our WSL method by introducing a window refinement method.  We also added additional experiments using CNN features, and their combination with FV features. Finally, we included experiments when training in a mixed supervision setting, where part of the images are weakly supervised and others are labeled with full bounding-box annotations.

\section{Related work}
\label{sec:related}

The majority of related work treats WSL for object
detection as a multiple instance learning
(MIL)~\cite{dietterich97ai} problem. Each image is
considered as a ``bag'' of examples given by tentative
object windows. Positive images are assumed to contain at
least one positive object instance window, while negative
images only contain negative windows. The object detector
is then obtained by alternating detector training, and
using the detector to select the most likely object
instances in positive images. 

In many MIL problems, \eg such as those for weakly
supervised face recognition
\cite{berg04cvpr,everingham09ivc}, the number of examples
per bag is limited to a few dozen at most.  In contrast,
there is a vast number of examples per bag in the case of
object detector training since the number of possible
object bounding boxes is quadratic in the number of image
pixels. Candidate window generation methods, \eg
\cite{alexe10cvpr,gu12eccv,uijlings13ijcv,zitnick14eccv}, can be
used to make MIL approaches to WSL for object localization manageable, and make it
possible to use powerful and computationally
expensive  object models. 

Although candidate window generation methods can
significantly reduce the search space per image, the
selection of windows across a large number of images is
inherently a challenging problem, where an iterative WSL
method can typically find only a local optimum depending
on the initial windows. Therefore, in 
this section, we first overview the
initialization methods proposed in the literature, and
then summarize the iterative WSL approaches.

\subsection{Initialization methods}

A number of different strategies to initialize the MIL
detector training have been proposed in the literature.  A
simple strategy,  \eg  taken in
\cite{pandey11iccv,russakovsky12eccv,kim09nips}, is to initialize
by taking large windows in positive images that (nearly)
cover the entire image. This strategy exploits the
inclusion structure of the MIL problem for object
detection. That is:  although large windows may contain a
significant amount of background features, they are likely
to include positive object instances.

Another strategy is to utilize a class-independent 
saliency measure that aims to predict whether a given
image region belongs to an object or not. For example,
Deselaers \etal \cite{deselaers12ijcv} generate candidate windows using
the objectness method \cite{alexe12pami} and assign
per-window weights using a saliency model trained on a
small set of non-target classes.
Siva \etal \cite{siva13cvpr} instead estimate an unsupervised
patch-level saliency map for a given image by measuring
the average similarity of each patch to the other patches
in a retrieved set of similar images. In each image, an initial window  is found by sampling from the corresponding
saliency map.

Alternatively, a class-specific initialization method can
be used. For example, Chum and Zisserman~\cite{chum07cvpr} select the
visual words that predominantly appear in the positive
training images and initialize WSL by finding the 
bounding box of these visual words in each image.
Siva and Xiang~\cite{siva11iccv} propose to initially select one of the
candidate windows sampled using the objectness method at
each image such that an objective function based on
intra-class and inter-class pairwise similarities is
maximized.  However, this formulation leads to a difficult
combinatorial optimization problem. Siva \etal \cite{siva12eccv}
propose a simplified approach where a candidate window is
selected for a given image such that the distance from the
selected window to its nearest neighbor among windows from
negative images is maximal.  Relying only on negative windows
not only avoids the difficult combinatorial optimization
problem, but also has the advantage that their labels are
certain, 
and there is a larger number of negative windows available
which makes the pairwise comparisons more robust. 

Shi \etal \cite{shi13iccv} propose to estimate a per-patch class
distribution by using an extended version of the Latent
Dirichlet Allocation (LDA) \cite{blei03jmlr} topic model.
Their approach assigns object class labels across
different object categories concurrently, which allows to
benefit from explaining-away effects, \ie an image region
cannot be identified as an instance for multiple
categories. The initial windows are then localized by
sampling from the saliency maps.

Song \etal \cite{song14icml} propose a graph-based
initialization method.  The main idea is to select a
subset of the candidate windows such that the nearest
neighbors of the selected windows correspond 
to the candidate windows in the positive images, rather
than the ones in the negative images. The approach is
formulated as a discriminative submodular cover problem on
the similarity graph of the 
windows. In a follow-up work, Song \etal
\cite{song14nips} extend this approach to find
multiple non-overlapping regions corresponding to object
parts.  The initial object windows are then generated by
finding frequent part configurations and their bounding
boxes.

\subsection{Iterative learning methods}
\label{sec:related_reloc}

Once the initial windows are localized, typically an
iterative learning approach is employed in order to
improve the initial localizations in the training images.

One of the early examples of WSL for object detector
training is proposed by Crandall and Huttenlocher \cite{crandall06eccv}.
In their work, object and part locations are treated as
latent variables in a probabilistic model.  These
variables are automatically inferred and utilized during
training using an  Expectation Maximization (EM)
algorithm. The main focus of their work, however, is
 on training a part-based object detector without
using manual part annotations, rather than training in
terms of image labels. Their approach is
evaluated on datasets containing images with
uncluttered backgrounds and little variance in terms of
object locations, which is an unrealistic testbed for
WSL of object detectors.

Several WSL methods aim to localize objects via selecting
a subset of candidate windows based on pairwise
similarities.  For example, Kim and Torralba
\cite{kim09nips} use a  link analysis based
clustering approach.  Chum and Zisserman \cite{chum07cvpr}
iteratively select windows and update the similarity
measure that is used to compare windows.  The window
selection is done by updating one image at a time such
that the average pairwise similarity across the positive
images is maximized. The similarity measure, which is
defined in terms of bag-of-word (BoW) descriptors~\cite{dance04eccv}, is updated by
selecting the visual words that predominantly appear in
the selected windows rather than the negative images.  

Deselaers \etal \cite{deselaers12ijcv} propose a CRF-based
model that jointly infers the object hypotheses across all
positive training images, by exploiting a fully-connected
graphical model that encourages visual similarity across
all selected object hypotheses.  Unlike the methods of
\cite{kim09nips} and \cite{chum07cvpr}, the CRF-based
model additionally utilize a  unary potential
function that scores candidate windows individually based on their 
window descriptors and objectness scores. The
parameters of the pairwise and unary potential functions
are updated, and the positive windows are selected in an
iterative fashion. Prest \etal \cite{prest12cvpr} extend
these ideas to weakly supervised detector training from
videos by extracting candidate spatio-temporal tubes based
on motion cues and by defining WSL potential functions
over tubes instead of windows.

Our window refinement method is inspired from the use of
an objectness model as a class-independent prior in
\cite{deselaers12ijcv}. While Deselaers~\etal\cite{deselaers12ijcv} use the objectness prior in all training iterations,
we update the coordinates of
the top-scoring final localizations, 
using the local greedy search procedure from
\cite{zitnick14eccv}.  In addition, instead of using the
objectness model in \cite{alexe12pami}, we use the
edge-driven objectness measure \cite{zitnick14eccv},
which evalutes the alignment between each window and 
the edges around it. 

Most recent work is predominantly based on iteratively 
selecting the highest scoring detections as the positive
training examples and training the detection models. 
We refer to this approach as {\em standard MIL}. 
Using this approach, an off-the-shelf detector can be trained
in a weakly supervised setting.
For example, Nguyen \etal \cite{nguyen09iccv} and
Blaschko \etal \cite{blaschko10nips} train the branch-and-bound
localization \cite{lampert09pami} based detectors over
BoW descriptors  in this manner. 
Blaschko \etal  also investigate 
the use of object-center annotations as an alternative WSL setting.

The DPM model \cite{felzenszalb10pami} has been utilized
with standard MIL based training approaches by a number of other WSL approaches, see \eg
\cite{siva11iccv,siva12eccv,shi13iccv,siva13cvpr,pandey11iccv}.
The majority of the works use the standard DPM training
procedure and differ in terms of their initialization
procedures. One exception is that Siva and Xiang \cite{siva11iccv}
propose a method to detect when the iterative training
procedure drifts to background regions. In addition,
Pandey and Lazebnik \cite{pandey11iccv} carefully study how to tune DPM
training for WSL purposes. They propose
to restrict each re-localization stage such that the
bounding boxes between two iterations must meet a minimum
overlap threshold, which avoids big fluctuations across
the iterations. Moreover, they propose a heuristic to
automatically crop windows with near-uniform backgrounds.

Russakovsky \etal \cite{russakovsky12eccv} use a similar approach based on
Locality-constrained Linear Coding
descriptors \cite{wang10cvpr} over the candidate
windows generated using the Selective Search method 
\cite{uijlings13ijcv}.  
They use a background
descriptor computed over features outside the window,
which helps to better localize the objects as compared to
only modeling the windows themselves. 

Song \etal \cite{song14icml} develop a smoothed version of
the standard MIL approach using Nesterov's smoothing
technique \cite{nesterov05mathprog}.  The main motivation
is to increase robustness against incorrectly selected
windows, particularly in early iterations, by training
with multiple windows per positive image.  
The candidate windows are generated using selective
search~\cite{uijlings13ijcv} and the window descriptors
are extracted using the CNN model of 
\cite{krizhevsky12nips}.

Bilen \etal \cite{bilen14bmvc} propose an alternative
smoothed version of standard MIL. Instead of
selecting the top scoring window in a positive image, they
propose to train over all windows that are weighted by a
soft-max function over the classification scores. In
addition, they utilize additional regularization terms
that aim to (i)~enforce that positive training windows and
their horizontal mirrors score similarly and,
(ii) avoid obtaining high classification scores for
multiple classes for a single window. They also utilize selective search candidate
windows \cite{uijlings13ijcv} and  CNN features \cite{krizhevsky12nips}.

Recently, Wang \etal \cite{wang14eccv} propose a two-step
method, which first groups selective search candidate
windows \cite{uijlings13ijcv} from the positive images of
a class into visual clusters and then chooses the most
discriminative cluster of windows. In the first step, the CNN features \cite{krizhevsky12nips} are clustered using probabilistic
latent semantic analysis (PLSA) 
\cite{hofmann01ml}. 
In the second
step, for each visual cluster, image descriptors are extracted from the CNN-based window descriptors of the windows associated with the cluster. Finally, one visual
cluster for each class is selected based on the image
classification performance of the corresponding image
descriptors.

Our approach is most related to that of Russakovsky \etal
~\cite{russakovsky12eccv}. We also rely on the selective
search windows \cite{uijlings13ijcv}, and use a similar
initialization strategy. A critical  difference from
\cite{russakovsky12eccv} and other WSL approaches based on
iterative detector training, however, is our multi-fold
MIL training procedure which we describe in the next
section.  Our multi-fold MIL approach is also related to
the work of Singh \etal \cite{singh12eccv} on unsupervised
vocabulary learning for image classification.  Starting
from an unsupervised clustering of local patches, they
iteratively train SVM classifiers on a subset of the data,
and evaluate it on another set to update the training data
from the second set.  

We note that avoiding poor local optima in training of
models with non-convex objectives is a
fundamental problem in machine learning, and there are many 
aspects of it.  For example,
curriculum learning (CL) \cite{bengio09icml}, which is a 
conceptual framework, suggests that training can be improved
by initializing a model with easy examples,
and then, gradually utilizing more complex ones.
Kumar \etal \cite{kumar10nips} propose a CL formulation 
for latent variable models
by considering the loss function as a measure
of example difficulty, which excludes low-scoring examples 
in early training iterations. 
Progressively increasing the latent search space
can also be interpreted as a CL approach
to avoid making unstable inferences in early iterations, see \eg \cite{russakovsky12eccv,bilen14ijcv}.
Although our work is related,  our focus is different in the sense that we target the problem of degenerate latent variable inference due to use of high-dimensional descriptors.

\section{Weakly supervised object localization}
\label{sec:method}

Below, we present our multi-fold MIL approach in \sect{training}
and window refinement method in \sect{refinement}, but
first briefly describe our FV and CNN based object
appearance descriptors. 

\subsection{Features and detection window representation}
\label{sec:windows}

In our experiments we rely on FV and CNN based
representations.  In either case, we use the selective
search method of Uijlings \etal~\cite{uijlings13ijcv}. It
generates a limited set of around 1,500 candidate windows
per image.  This speeds-up detector training and
evaluation, while filtering out the most implausible
object locations.  

The FV-based representation is based on our previous
work~\cite{cinbis13iccv} for fully supervised detection.  In
particular, we aggregate local SIFT descriptors into an FV
representation to which we apply $\ell_2$ and power
normalization~\cite{sanchez13ijcv}. We concatenate the FV
computed over the full detection window, and  16 FVs
computed over the cells in a $4\times 4$ grid over the
window, inspired by the spatial pyramid representation of
Lazebnik \etal~\cite{lazebnik06cvpr}.  Using PCA to
project the SIFTs to 64 dimensions, and a mixture of
Gaussians (MoG) of 64 components, this yields a
descriptor of 140,352 dimensions. We reduce the memory
footprint, and speed up our iterative training procedure,
by using the PQ and Blosc feature
compression~\cite{alted10cse,jegou11pami2}.

Similar to Russakovsky \etal \cite{russakovsky12eccv}, we
add contextual information from the part of the image not
covered by the window. Full-image descriptors, or image
classification scores, are commonly used for fully
supervised object detection, see \eg
\cite{cinbis13iccv,song2011cvpr}.  For WSL, however, it is
important to use the complement of the object window
rather than the full image, to ensure that the context
descriptor also depends on the window location. This
prevents learning degenerate detection models, since otherwise the context descriptor can be used
to perfectly separate the training images regardless of
the object localization.

To enhance the effectiveness of the context descriptor we
propose a ``contrastive'' version, defined as the
difference between the background FV $\bm x_b$ and the
$1\times 1$ foreground FV $\bm x_f$.  Since we use linear
classifiers, the contribution to the window score of this
descriptor, given by $\bm w^\top(\bm x_b-\bm x_f)$, can be
decomposed as a sum of a foreground and a background
score: $\bm w^\top \bm x_b$ and   $-\bm w^\top \bm x_f$
respectively.  Because the foreground and background
descriptor have the same weight vector, up to a sign flip,
we effectively  force features to either score positively
on the foreground and negatively on the background, or
\emph{vice-versa} within the contrastive descriptor. This prevents the detector to score the
same features positively on both the foreground and the
background.

To ensure that we have enough SIFT descriptors for the
background FV, we filter the detection windows to respect
a margin of at least 4\% from the image border, \ie for a
$100\times 100$ pixel image, windows closer than 4 pixels
to the image border are suppressed.  This filtering step
removes about half of the windows.  We initialize the MIL
training with the window that covers the image, up to a
4\% margin, so that all instances are captured by the
initial windows.

We extract the CNN features using the CNN architecture of
Krizhevsky \etal \cite{krizhevsky12nips}.  We utilize the
first seven layers of the CNN model, which consists of
five convolutional and two fully-connected layers.  The
CNN model is pre-trained on the ImageNet ILSVRC 2012
dataset using the Caffe framework \cite{jia14caffe}.
Following Girshick \etal \cite{girshick14cvpr}, we crop
and resize the mean-subtracted regions corresponding to the
candidate windows to images of size $224\times224$, as
required by the CNN model. Finally, we apply $\ell_2$
normalization to the resulting 4096 dimensional 
descriptors.

An important advantage of the CNN features is that some of
the feature dimensions correspond to higher level image
structures, such as certain animal faces and bodies
\cite{girshick14cvpr}, which can simplify the WSL problem.
Our experimental result show that the CNN features perform better
than the FV features, but that they are complementary since best performance is obtained when combining both features.

\begin{figure}
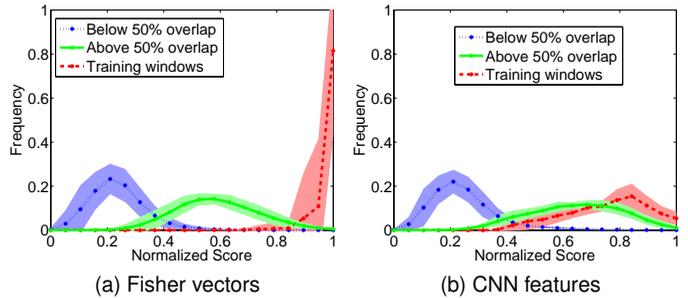

\def\myfig#1{\includegraphics[width=0.5\linewidth]{#1}}
\begin{center}
\subfloat[Fisher vectors\label{fig:scorehist:FV}]{%
    \myfig{journal/scorehistogram/2007+standardMIL}
}
\subfloat[CNN features\label{fig:scorehist:CNN}]{%
    \myfig{journal/scorehistogram/journal/2007+standardMIL+CNNonly}
}
\end{center}
\vspace{-1mm}
\caption{Distribution of the window scores in the positive training images after the
fifth iteration of standard MIL training on VOC 2007 for FVs (left) and CNNs (right). 
For each figure, the right-most curve  
corresponds to the windows  chosen in the most recent re-localization step
and used for training the detector.
The curve in the middle corresponds to the other windows
that overlap more than 50\% with the
training windows. Similarly, the left-most curve
corresponds to the windows that overlap less than 50\%.
Each curve is obtained by averaging all per-class score distributions. 
The surrounding regions show the standard deviation.}
\label{fig:scorehist}
\vspace{-1mm}
\end{figure}

\subsection{Weakly supervised object detector training}
\label{sec:training}

The dominant method for weakly supervised training of
object detectors is the standard MIL approach, which is
based on iterating between the training and the
re-localization stages, as described in \sect{related_reloc}.
Note that in this approach, the detector used for
re-localization in positive images is trained using
positive samples that are extracted from the very same
images.  Therefore, there is a bias towards re-localizing
on the same windows; in particular when high capacity
classifiers are used which are likely to separate the
detector's training data.  For example, when a nearest
neighbor classifier is used the re-localization will be
degenerate and not move away from its initialization,
since the same window will be found as its nearest
neighbor.

The same phenomenon occurs when using powerful and
high-dimensional image representations to train linear
classifiers. We illustrate this in \fig{scorehist}, which
shows the distribution of the window scores in a typical
standard MIL iteration.
We observe that the windows used in SVM training score
significantly higher than the other ones, including those
with a significant spatial overlap with the most recent training
windows, especially when the high-dimensional FV
descriptors are used.

\begin{figure}
\def\myfig#1{\includegraphics[width=0.34\linewidth]{dotprodhist/journal_withNegatives/StrechMinToMinusOne/#1}}
\begin{center}
\subfloat[All pairs.]{
        {\centering
            \begin{tabular}{c}
                \myfig{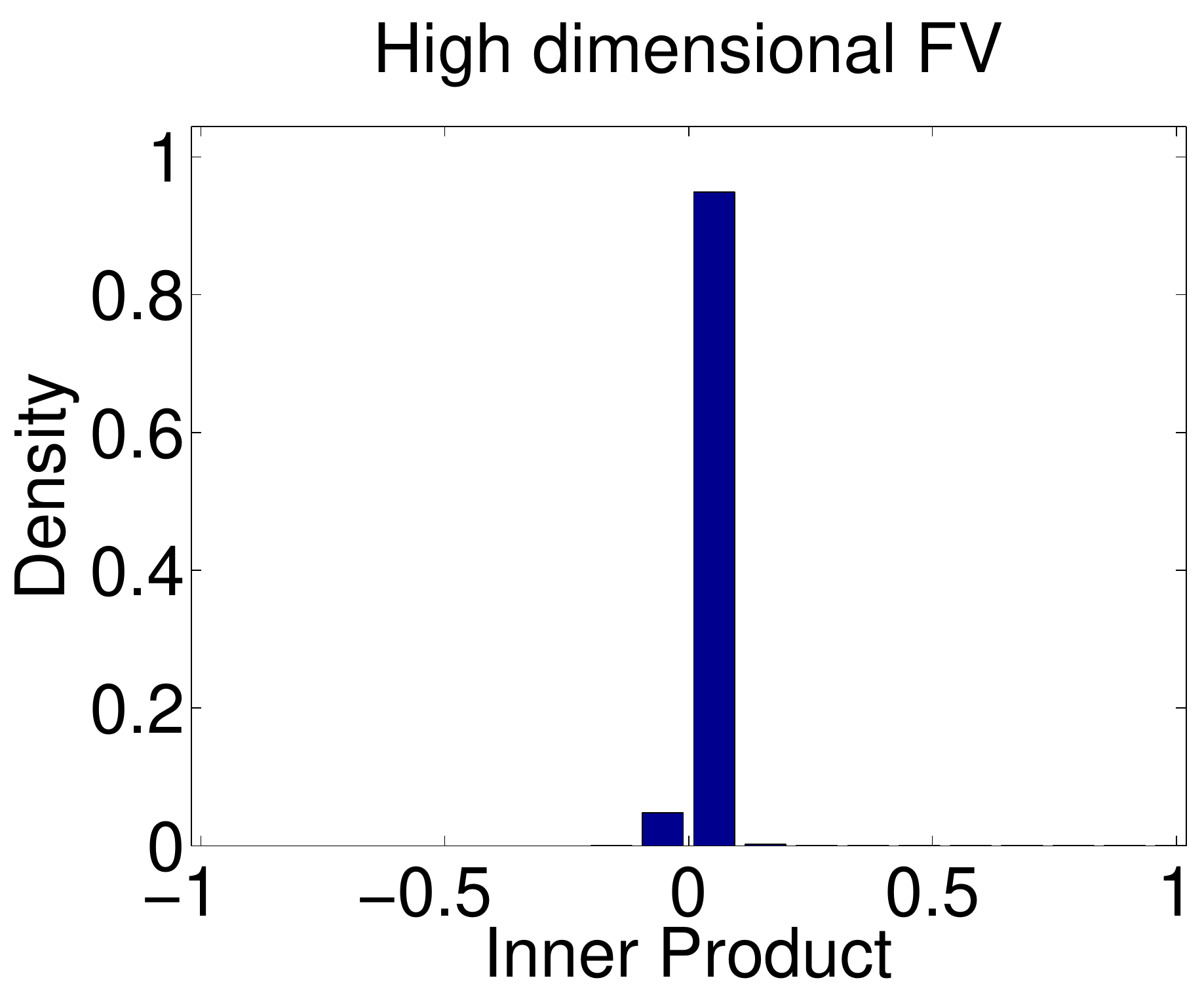} \\[1pt]
                \myfig{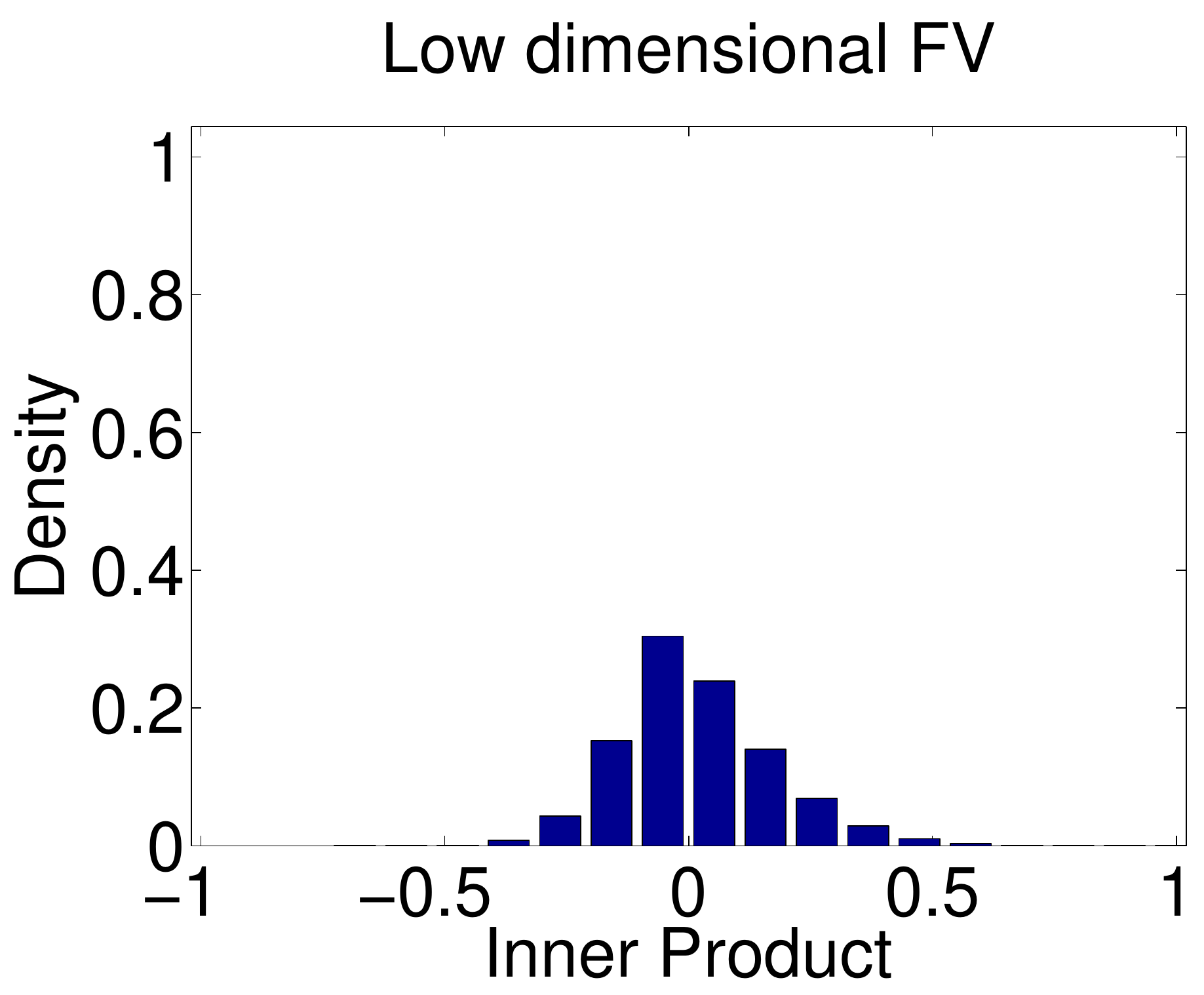}  \\[1pt]
                \myfig{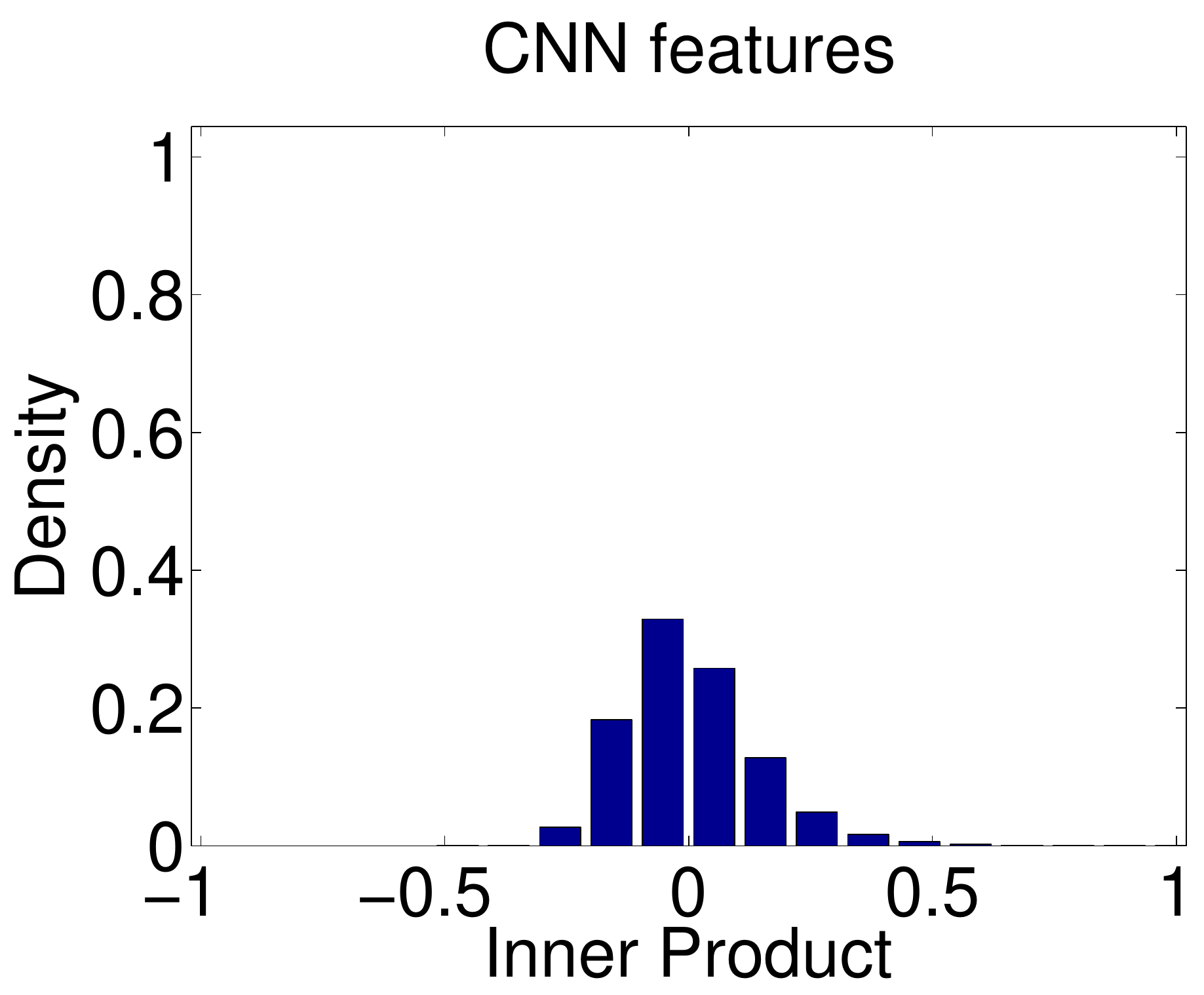}  
            \end{tabular}
        }
        \vspace{-1mm}
        \label{fig:dotprodhist:allpairs}
    }
\hspace{1mm}
\subfloat[Within-image pairs.]{
        {\centering
            \begin{tabular}{c}
                \myfig{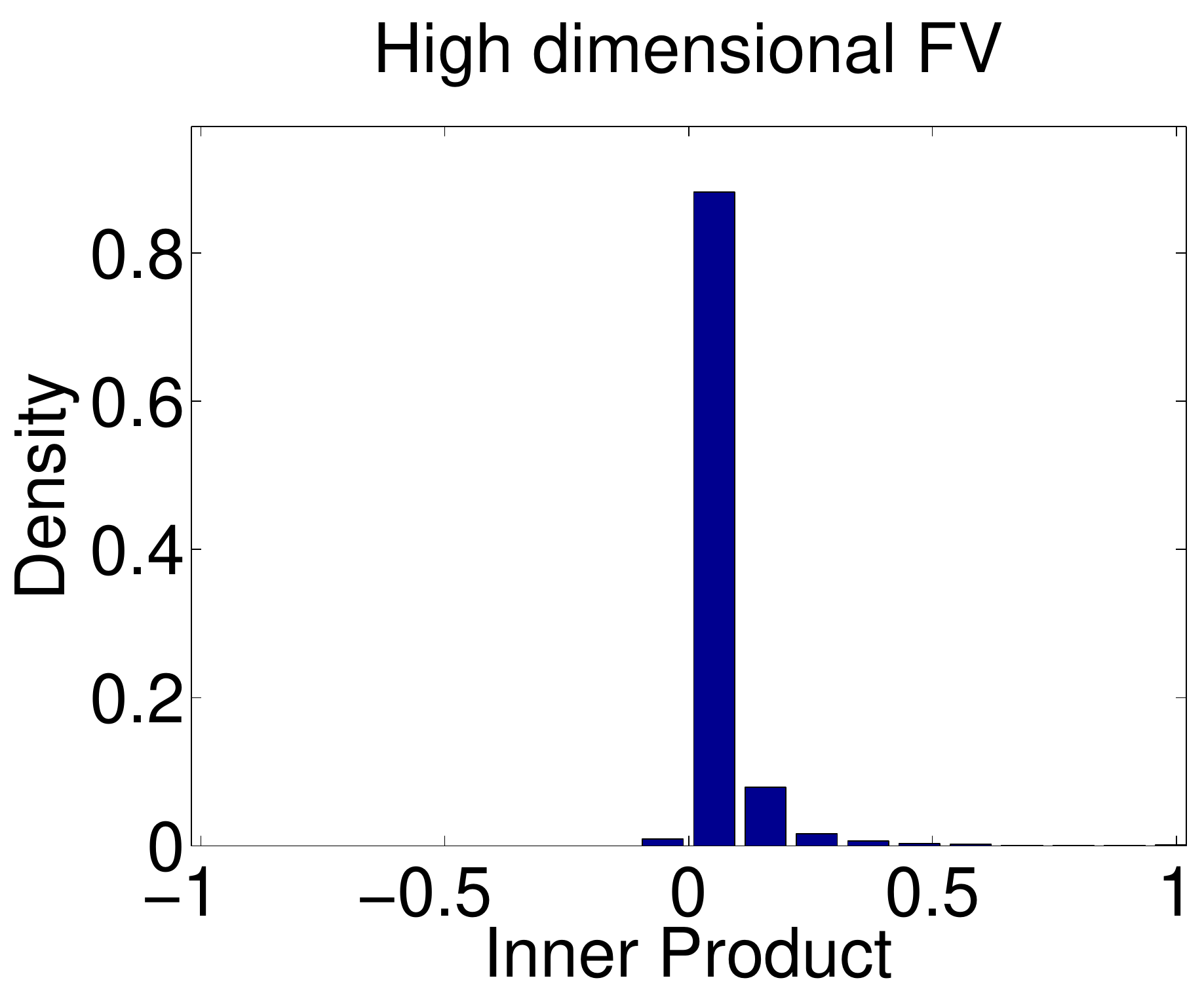} \\[1pt]
                \myfig{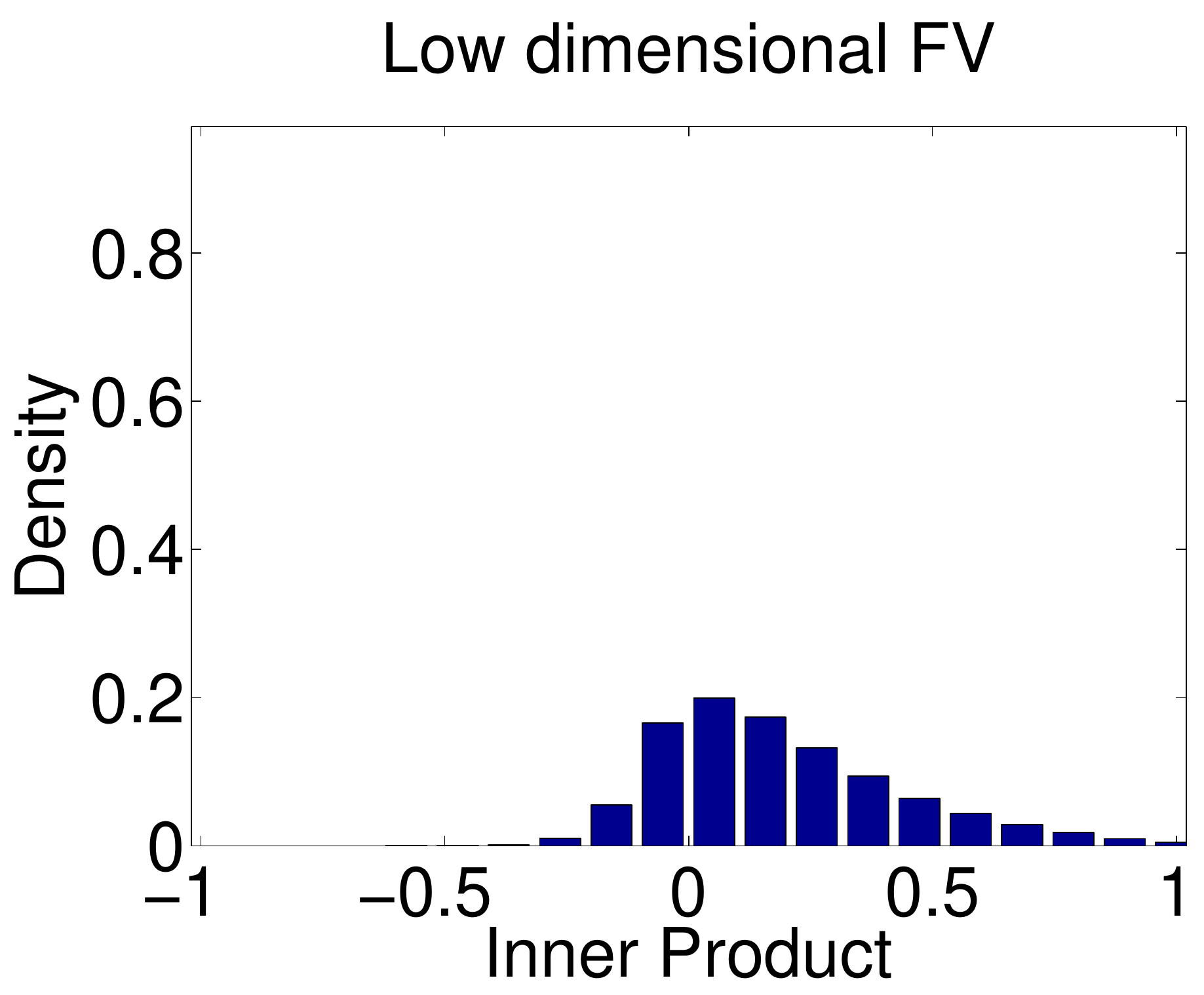}  \\[1pt]
                \myfig{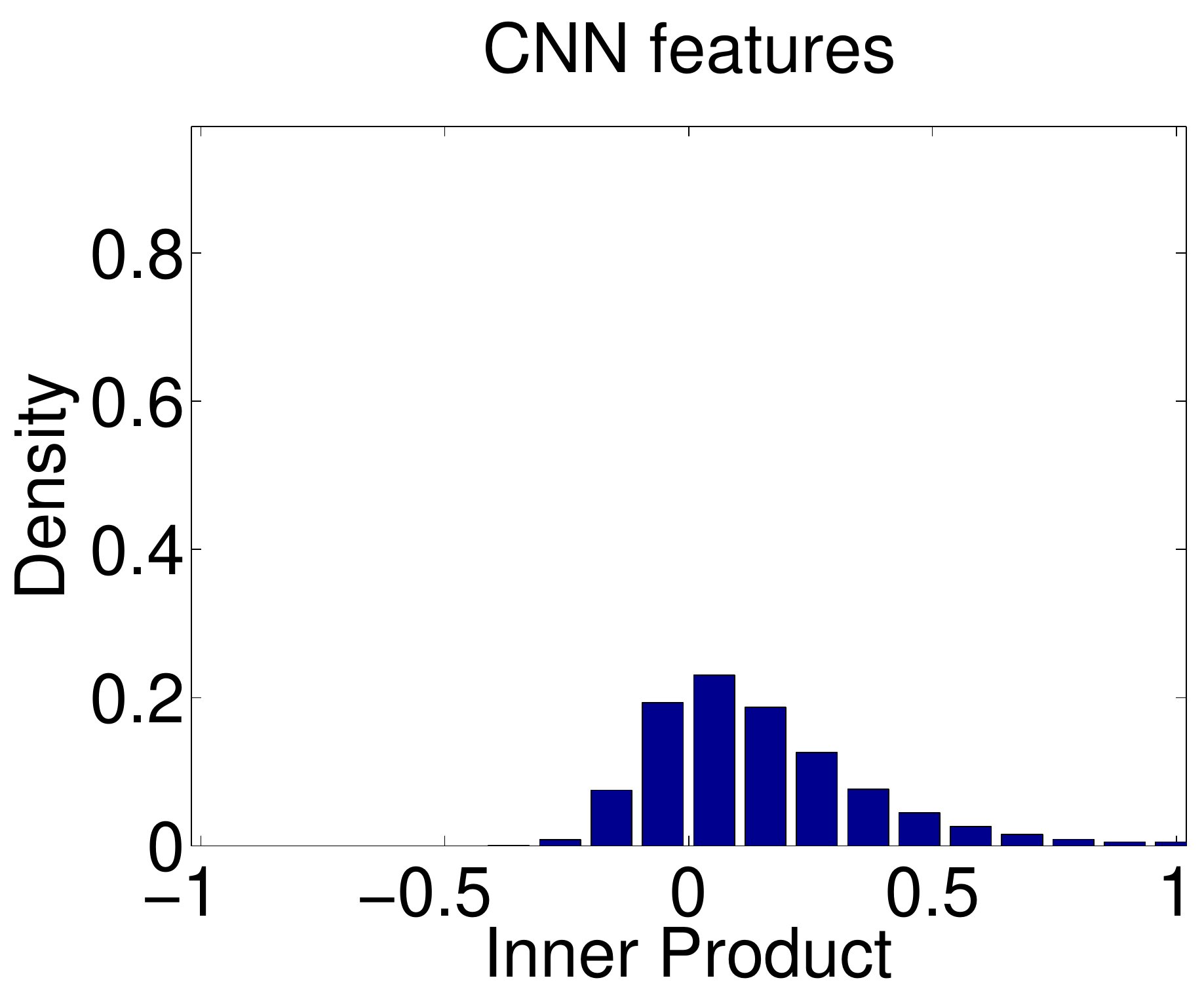}  
            \end{tabular}
        }
        \vspace{-1mm}
        \label{fig:dotprodhist:inimg}
    }
\end{center}
\vspace{-3mm}
\caption{Distribution of inner products, scaled to the
interval [-1 +1], of pairs of 25,000 windows sampled from 250 images using our
high-dimensional FV (top), a low-dimensional FV (middle), and CNN features
(bottom). (a) uses all window pairs and (b) uses only
within-image pairs, which are more likely to be similar.
}
\label{fig:dotprodhist}
\vspace{-1mm}
\end{figure}

As a result, standard MIL typically results in degenerate
re-localization. This problem is related to the
dimensionality of the window descriptors.  We illustrate
this in \fig{dotprodhist}, where we show the distribution
of inner products between the descriptors of different
windows.  In \fig{dotprodhist:allpairs}, we use random
window pairs within and across images. In
\fig{dotprodhist:inimg}, we use only within-image pairs,
which are more likely to be similar, and therefore the
histograms models are shifted slightly to larger values.
We show the distributions using both our 140,352
dimensional FVs, 516 dimensional FVs obtained using 4
Gaussians without spatial grid, and 4096 dimensional
CNN-based descriptors.\footnote{To make the histograms comparable, we make all descriptors zero mean, before $\ell_2$ normalization, and computing the inner products.}
Unlike in the case of
low-dimensional FVs or CNN-based descriptors, almost all
window descriptors are near orthogonal in the
high-dimensional FV case even when we use within-image
pairs only. Also, recall that the weight vector of a
standard linear SVM classifier can be written as a linear
combination of training samples, $\bm w=\sum_i \alpha_i
\bm x_i$.  Therefore, the training windows are likely to
score significantly higher than the other windows in
positive images in the high-dimensional case, resulting in
degenerate re-localization behavior. In
\sect{experiments}, we verify this hypothesis
experimentally by comparing the localization behavior
using the low-dimensional \vs the high-dimensional
descriptors. 

Note that increasing regularization weight in SVM training does not remedy this problem. The
$\ell_2$ regularization term with weight $\lambda$
restricts the linear combination weights such that
$|\alpha_i|\leq{1/\lambda}$.  Therefore, although we can
reduce the influence of individual training samples via
regularization, the resulting classifier remains biased
towards the training windows since the classifier is a
linear combination of the window descriptors. In
\sect{experiments}, we verify this hypothesis 
by evaluating the regularization weight's effect on the
localization performance.

To address this issue---without sacrificing the descriptor 
dimensionality, which would limit its descriptive power---we propose
to train the detector using a multi-fold 
procedure, reminiscent of cross-validation, within the MIL
iterations.  We divide the positive training images into
$K$ disjoint folds, and re-localize the images in each
fold using a detector trained using windows from positive
images in the other folds.  In this manner the
re-localization detectors never use training windows from
the images to which they are applied.  Once
re-localization is performed in all positive training
images, we train another detector using all selected
windows. This detector is used for hard-negative mining on
negative training images, and returned as the final
detector.

\begin{algorithm}[t]
\caption{--- Multi-fold weakly supervised training}
{
\begin{enumerate}
\item Initialization: positive and negative examples are set to entire images up to a 4\%  border
\item For iteration $t=1$ to $T$
    \begin{enumerate}[itemsep=0mm,topsep=-.5mm]
    \item Divide positive images randomly into $K$ folds
    \item For $k=1$ to $K$ 
        \begin{enumerate}[itemsep=0mm,topsep=-.5mm]
        \item Train using positive examples in all folds but $k$, and all negative examples
	\item Re-localize positives by selecting the top scoring window in each image of fold $k$ using this detector
        \end{enumerate}
    \item Train detector using re-localized positives and all negatives
    \item Add new negative windows by hard-negative mining 
    \end{enumerate}
\item Return final detector and object windows in train data
\end{enumerate}}
\label{algo:kfold}
\end{algorithm}

We summarize our {\em multi-fold MIL} training procedure
in \algo{kfold}.  The standard MIL algorithm that does not
use multi-fold training does not execute steps 2(a) and
2(b), and re-localizes based on the detector learned in
step 2(c).

The number of folds used in our multi-fold MIL training
procedure should be set to strike a good trade-off between
two competing factors. On the one hand, using more folds
increases the number of training samples  per fold, and is
therefore likely to improve re-localization performance.
On the other hand, using more folds 
increases the computational cost. We
experimentally analyze this trade-off in \sect{experiments}.

\subsection{Window refinement}
\label{sec:refinement}

We now explain our window refinement method. It updates the
localizations obtained by the last multi-fold MIL iteration. The final
detector is, then, re-trained based on these refinements.

An inherent difficulty for weakly supervised object localization is
that WSL labels only permit to determine the 
most repeatable and discriminative patterns for each
class.  Therefore, even though the windows found by WSL
are likely to overlap with target object instances, it can not be 
ensured that they will delineate object boundaries. 

To better take into account object boundaries, we use the
edge-driven {\em objectness} measure of Zitnick and
Dollar~\cite{zitnick14eccv}. The main idea in~\cite{zitnick14eccv} is to score a
given window based on the number of contours that are
fully contained inside the window, with an increased weight
on near-boundary edge pixels. Thus, windows that tightly
enclose long contours are scored highly, whereas those
with predominantly straddling contours are penalized. Additionally, in
order to reduce the effect of marginal misalignments, the coordinates
of a given window are updated using a greedy local search
procedure that aims to increase the objectness score.
In~\cite{zitnick14eccv}, the objectness measure is used
for generating object proposals. For this purpose, a set
of initial windows are first generated using a sliding window
mechanism, and then, updated and scored using
the local search procedure. The final windows
are obtained by applying a non-maxima suppression
procedure. 

\def\myfig#1{\adjustimage{width=36mm,valign=m,margin=0mm 0.4mm}{ws_script_xpmainoutputs/voc07_fv+cnn/#1}}
\begin{figure}
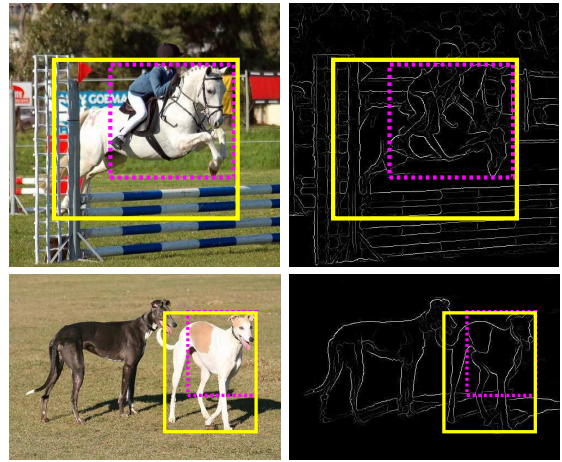

\vspace{-1mm}
\begin{center}
{\addtolength{\tabcolsep}{-2pt}
\begin{tabular}{c}
\myfig{horse/003779_im} \myfig{horse/003779_edges} \\
\myfig{dog/009870_im} \myfig{dog/009870_edges} \\ 
\end{tabular}
}
\vspace{-1mm}
\end{center}
\caption{Illustration of our window refinement. Dashed boxes (pink) show the localization before refinement, and the 
solid boxes (yellow) show the result of the window refinement method. The images on the right show the edge
maps that are used to compute the objectness measure.}
\vspace{-2mm}
\label{fig:refinement}
\end{figure}

\def\myStd{\rotatebox[origin=c]{90}{{\small standard}}}
\def\myMulti{\rotatebox[origin=c]{90}{{\small multi-fold}}}
\def\myfig#1{\adjustimage{width=26.5mm,height=20mm,valign=m,margin=0mm 0.4mm}{iteration_images/selected/voc2007/#1}}
\def\myfigStd#1{\myStd      & \myfig{#1-0-standard} \myfig{#1-1-standard} \myfig{#1-4-standard} \myfig{#1-8-standard} \myfig{#1-11-standard} \\}
\def\myfigMulti#1{\myMulti  & \myfig{#1-0-kfold}    \myfig{#1-1-kfold}    \myfig{#1-4-kfold}    \myfig{#1-8-kfold}    \myfig{#1-11-kfold} \\}
\begin{figure*}
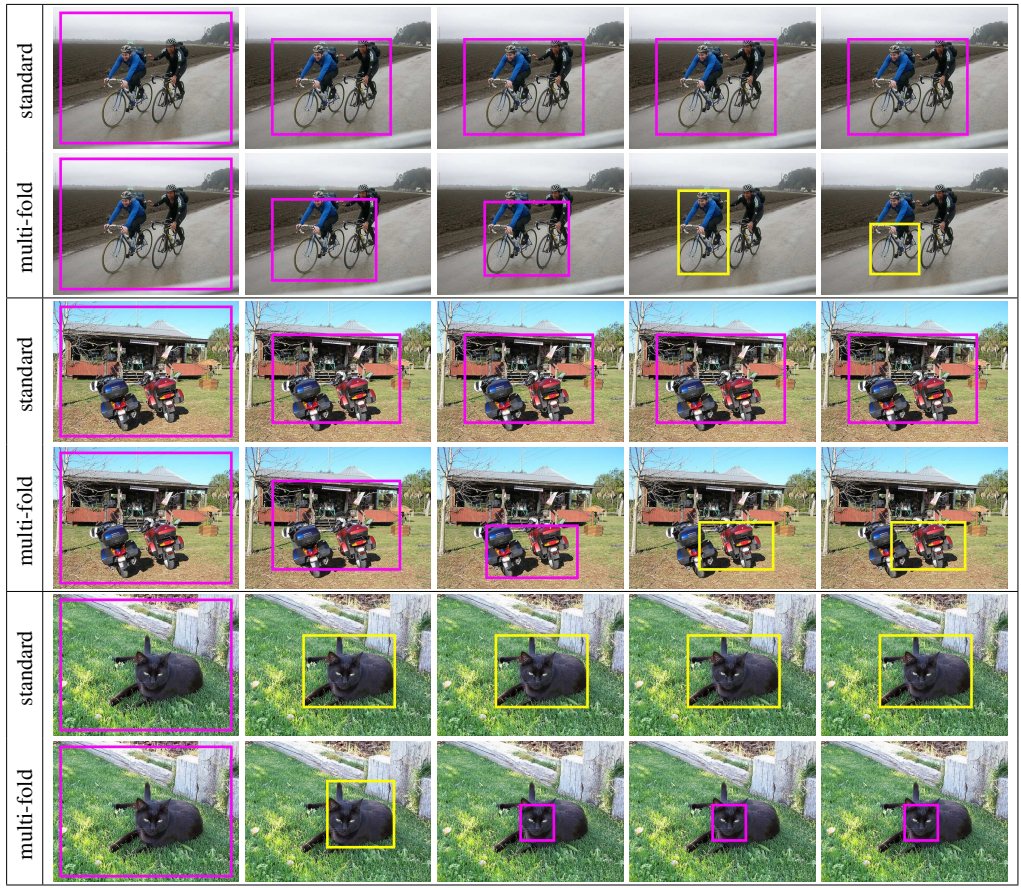

\begin{center}
\resizebox{.75\textwidth}{!}{ 
{\addtolength{\tabcolsep}{-2pt}
\begin{tabular}{|l|c|}
\hline
\myfigStd{bicycle/007571}
\myfigMulti{bicycle/007571}
\hline
\myfigStd{motorbike/000822}
\myfigMulti{motorbike/000822}
\hline
\myfigStd{cat/002480}
\myfigMulti{cat/002480}
\hline
\end{tabular}
}}
\end{center}
\caption{Re-localization using standard and multi-fold MIL for images of the
classes {\em bicycle, motorbike,} and {\em cat} from initialization (left) to the final
localization (right) and three intermediate iterations based on FV (F+C) descriptors.
Correct localizations are shown in yellow, incorrect ones
in pink.  }
\label{fig:imagesiter:fv}
\end{figure*}

We instead use the edge-driven objectness measure to improve our WSL
outputs. For this
purpose, we combine the objectness measure with the
classification scores given by multi-fold MIL.
More specifically, we first utilize the local search
procedure in order to update and score the 
candidate detection windows based on the objectness measure,
without updating
the classification scores. 
To make the classification and
objectness scores comparable, we scale each score channel
to the range $[0,1]$ for all windows in the positive
training images.  We, then, combine linearly the
classification and objectness scores with equal
weights, and select the top detection in each image with
respect to this combined score. In order to avoid selecting the
windows irrelevant for the target class, but with a high objectness
score, we restrict the search space to the top-$N$ windows per image
in terms of the classification score. While we use $N=10$ in all our experiments,
we have empirically observed that the refinement method significantly
improves the localization results for  $N$ ranging from 1 to 50. 
The improvement is comparable for  $N\geq 5$. 

In \fig{refinement}, we show two example images for the
classes {\em horse} and {\em dog} in the left column,
together with the corresponding edge maps in the right
column. In these images, the dashed (pink) boxes show the
output of multi-fold MIL training and the solid (yellow)
boxes show the outputs of the window refinement procedure.
Even though the initial windows are located on the object
instances, they are evaluated as incorrect due to the low
overlap ratios with the ground-truth ones. The edge
maps show that many contours, \ie most object contours, straddle
the initial window boundaries. In contrast, the corrected
windows have higher percentages of fully contained contours, \ie the contours relevant for the objects.

{\addtolength{\tabcolsep}{-3.5pt}
\begin{table*}
\caption{Weakly supervised learning using FV and CNN features, measured in terms of correct localization (CorLoc) measure on VOC 2007 training set. We compare foreground (F), background (B) and contrastive background (C) FVs. Contrastive background is used in the FV+CNN combination.}
\label{tab:voc07_eval_corloc}
\begin{center}
\begin{tabular}{|l|cccccccccccccccccccc|c|}
\cline{2-22}
\multicolumn{1}{l|}{} & aero          & bicy                    & bird                & boa           & bot                   & bus                     & car           & cat           & cha                   & cow           & dtab          & dog           & hors              & mbik          & pers                  & plnt                & she           & sofa          & trai                  & tv                    & Av.\tabularnewline
\hline
 & \multicolumn{20}{c}{standard MIL}   & \tabularnewline
\hline
FV: F                 & 46.2          & 32.2                    & 32.0                & 24.1          & 4.0                   & 45.1                    & 51.5          & 37.6          & 6.8                   & 24.3          & 14.3          & 43.0          & 36.2              & 52.7          & 19.3                  & 9.3                 & 20.3          & 24.5          & 45.1                  & 14.2                  & 29.1 \tabularnewline
\hline
FV: F+B               & 50.3          & 32.2                    & 32.4                & 24.8          & 4.0                   & 45.1                    & 52.2          & {{{{41.1}}}}  & 6.8                   & 25.2          & 14.3          & {{{{44.1}}}}  & 38.2              & 53.7          & 20.5                  & 9.3                 & 20.3          & 24.5          & 43.4                  & 14.2                  & 29.8 \tabularnewline
\hline
FV: F+C               & 48.6          & 32.8                    & 30.9                & 25.5          & 4.0                   & 43.4                    & 52.2          & 40.6          & 6.8                   & 27.2          & 14.3          & 43.7          & 38.6              & 52.7          & 20.0                  & 8.8                 & 20.3          & 24.5          & 45.1                  & 14.7                  & 29.7 \tabularnewline
\hline
CNN                   & 54.3          & 55.6                    & 49.5                & {{31.7}}      & 15.9                  & 61.5                    & {{72.2}}      & 33.2          & 16.5                  & 43.7          & 22.4          & 34.8          & {{\textbf{58.5}}} & 64.4          & 25.1                  & 31.9                & 36.2          & {{34.0}}      & 52.2                  & 31.5                  & 41.2 \\
\hline
FV+CNN                & 49.1          & 36.1                    & 38.9                & 30.3          & 5.1                   & 49.2                    & 62.4          & \textbf{47.5} & 6.8                   & 35.0          & 18.4          & \textbf{44.8} & 45.4              & 54.3          & 29.3                  & 13.2                & 26.1          & 29.2          & 48.7                  & 18.8                  & 34.4 \\
\hline
 & \multicolumn{20}{c}{multi-fold MIL} & \tabularnewline
\hline
FV: F                 & 48.0          & 55.6                    & 25.8                & 4.1           & 6.3                   & 53.3                    & 68.3          & 23.3          & 8.8                   & 57.3          & 4.1           & 27.6          & 52.7              & {{{{66.0}}}}  & 33.2                  & 15.4                & 55.1          & 14.2          & 49.6                  & {{{{\textbf{62.4}}}}} & 36.5 \tabularnewline
\hline
FV: F+B               & 55.5          & 56.1                    & 21.8                & 27.6          & 4.5                   & 51.6                    & 66.5          & 19.3          & 8.4                   & {59.2}        & 2.0           & 26.2          & 56.0              & 64.9          & 35.5                  & 20.9                & {{{{58.0}}}}  & 10.4          & {{{{\textbf{56.6}}}}} & 59.4                  & 38.0 \tabularnewline
\hline
FV: F+C               & {{{{56.6}}}}  & 58.3                    & 28.4                & 20.7          & 6.8                   & 54.9                    & 69.1          & 20.8          & 9.2                   & 50.5          & 10.2          & 29.0          & {{58.0}}          & 64.9          & {{{{\textbf{36.7}}}}} & 18.7                & 56.5          & 13.2          & 54.9                  & 59.4                  & 38.8 \tabularnewline
\hline
CNN                   & 53.2          & {{{{{\textbf{66.7}}}}}} & {{{\textbf{51.3}}}} & {{{31.7}}}    & {{{19.3}}}            & {{{{{\textbf{70.5}}}}}} & 72.0          & 23.3          & {{{24.9}}}            & {{{62.1}}}    & {{{32.7}}}    & 28.0          & 54.6              & 64.9          & 22.1                  & {{{\textbf{39.0}}}} & 55.1          & {33.0}        & 54.9                  & 40.1                  & {{{45.0}}} \\
\hline
FV+CNN                & \textbf{57.2} & 62.2                    & 50.9                & \textbf{37.9} & {{{{\textbf{23.9}}}}} & 64.8                    & \textbf{74.4} & 24.8          & {{{{\textbf{29.7}}}}} & \textbf{64.1} & \textbf{40.8} & 37.3          & 55.6              & \textbf{68.1} & 25.5                  & {{38.5}}            & \textbf{65.2} & \textbf{35.8} & \textbf{56.6}         & 33.5                  & \textbf{47.3} \\
\hline
\end{tabular}
\end{center}
\end{table*}
}

{\addtolength{\tabcolsep}{-3.5pt}
\begin{table*}
\caption{Weakly supervised learning using FV and CNN features, measured in terms of average precision (AP) measure on VOC 2007 test set. We compare foreground (F), background (B) and contrastive background (C) FVs. Contrastive background is used in the FV+CNN combination.}
\label{tab:voc07_eval_ap}
\begin{center}
\begin{tabular}{|l|cccccccccccccccccccc|c|}
\cline{2-22}
\multicolumn{1}{l|}{} & aero          & bicy            & bird              & boa           & bot                     & bus           & car                     & cat           & cha                     & cow               & dtab          & dog           & hors            & mbik          & pers                  & plnt            & she               & sofa          & trai          & tv                    & Av.\tabularnewline
\hline
 & \multicolumn{20}{c}{standard MIL}   & \tabularnewline
\hline
FV: F                & 25.4          & 31.9            & 5.6               & 2.3           & 0.2                     & 27.9          & 35.4                    & 20.6          & 0.5                     & 6.8               & 4.9           & 14.0          & 17.0            & 35.2          & 7.1                   & 6.2             & 5.8               & 5.1           & 20.7          & 8.1                   & 14.0 \tabularnewline
\hline
FV: F+B              & 28.8          & 30.7            & 10.5              & 6.6           & 0.3                     & 30.1          & 36.2                    & {{{{22.7}}}}  & 0.9                     & 7.2               & 3.4           & 16.3          & 22.3            & 35.5          & 7.7                   & 9.2             & 7.5               & 3.9           & 26.2          & 6.5                   & 15.6 \tabularnewline
\hline
FV: F+C              & 26.1          & 31.6            & 8.3               & 5.3           & 1.3                     & 31.1          & 36.9                    & {{{{22.7}}}}  & 0.7                     & 7.7               & 2.1           & 16.6          & 24.5            & 36.7          & 7.7                   & 4.7             & 4.2               & 4.5           & 30.0          & 7.5                   & 15.5 \tabularnewline
\hline
CNN                   & 34.2          & 39.9            & 26.5              & 11.7          & 7.0                     & 38.0          & {45.6}                  & 19.6          & 6.2                     & 25.5              & 5.3           & {18.8}        & {\textbf{34.2}} & {42.3}        & 15.6                  & {\textbf{20.0}} & 18.6              & {23.5}        & {37.0}        & 15.8                  & 24.3 \\
\hline
FV+CNN                & 36.4          & 31.7            & 23.9              & 11.7          & 1.5                     & 37.8          & 40.4                    & \textbf{29.4} & 1.1                     & 17.1              & 5.1           & \textbf{29.0} & 32.3            & 40.9          & 15.2                  & 8.2             & 14.3              & 19.7          & 36.9          & 8.2                   & 22.0 \\
\hline
 & \multicolumn{20}{c}{multi-fold MIL} & \tabularnewline
\hline
FV: F                & 29.4          & 37.8            & 7.3               & 0.5           & 1.1                     & 33.2          & 41.0                    & 14.3          & 1.0                     & 21.9              & 9.2           & 9.4           & 29.1            & 37.3          & 15.5                  & 9.8             & {{27.9}}          & 4.7           & 29.4          & 40.4                  & 20.0 \tabularnewline
\hline
FV: F+B              & {{{{36.7}}}}  & 39.2            & 8.2               & 10.4          & 1.9                     & 31.4          & 40.4                    & 15.7          & 1.6                     & 22.6              & 5.8           & 7.4           & 29.1            & 40.9          & 18.9                  & 10.4            & 27.3              & 2.9           & 30.1          & 38.2                  & 21.0 \tabularnewline
\hline
FV: F+C              & 35.8          & 40.6            & 8.1               & 7.6           & 3.1                     & 35.9          & 41.8                    & 16.8          & 1.4                     & 23.0              & 4.9           & 14.1          & 31.9            & {{{41.9}}}    & {{{{\textbf{19.3}}}}} & 11.1            & 27.6              & 12.1          & 31.0          & {{{{\textbf{40.6}}}}} & 22.4 \tabularnewline
\hline
CNN                   & 32.1          & {{46.9}}        & {{\textbf{28.4}}} & {{12.0}}      & {{9.6}}                 & {39.4}        & 45.5                    & 16.2          & {{14.8}}                & {{\textbf{33.1}}} & {{11.6}}      & 14.0          & 31.2            & 39.3          & 13.1                  & {19.7}          & {{\textbf{30.5}}} & {23.4}        & {37.0}        & 19.6                  & {{25.9}} \\
\hline
FV+CNN                & \textbf{38.1} & {\textbf{47.6}} & 28.2              & \textbf{13.9} & {{{{{\textbf{13.2}}}}}} & \textbf{45.2} & {{{{{\textbf{48.0}}}}}} & 19.3          & {{{{{\textbf{17.1}}}}}} & 27.7              & \textbf{17.3} & 19.0          & 30.1            & \textbf{45.4} & 13.5                  & 17.0            & 28.8              & \textbf{24.8} & \textbf{38.2} & 15.0                  & \textbf{27.4} \\
\hline
\end{tabular}
\end{center}
\end{table*}
}

\begin{figure*}
\begin{center}
\includegraphics[height=.18\linewidth]{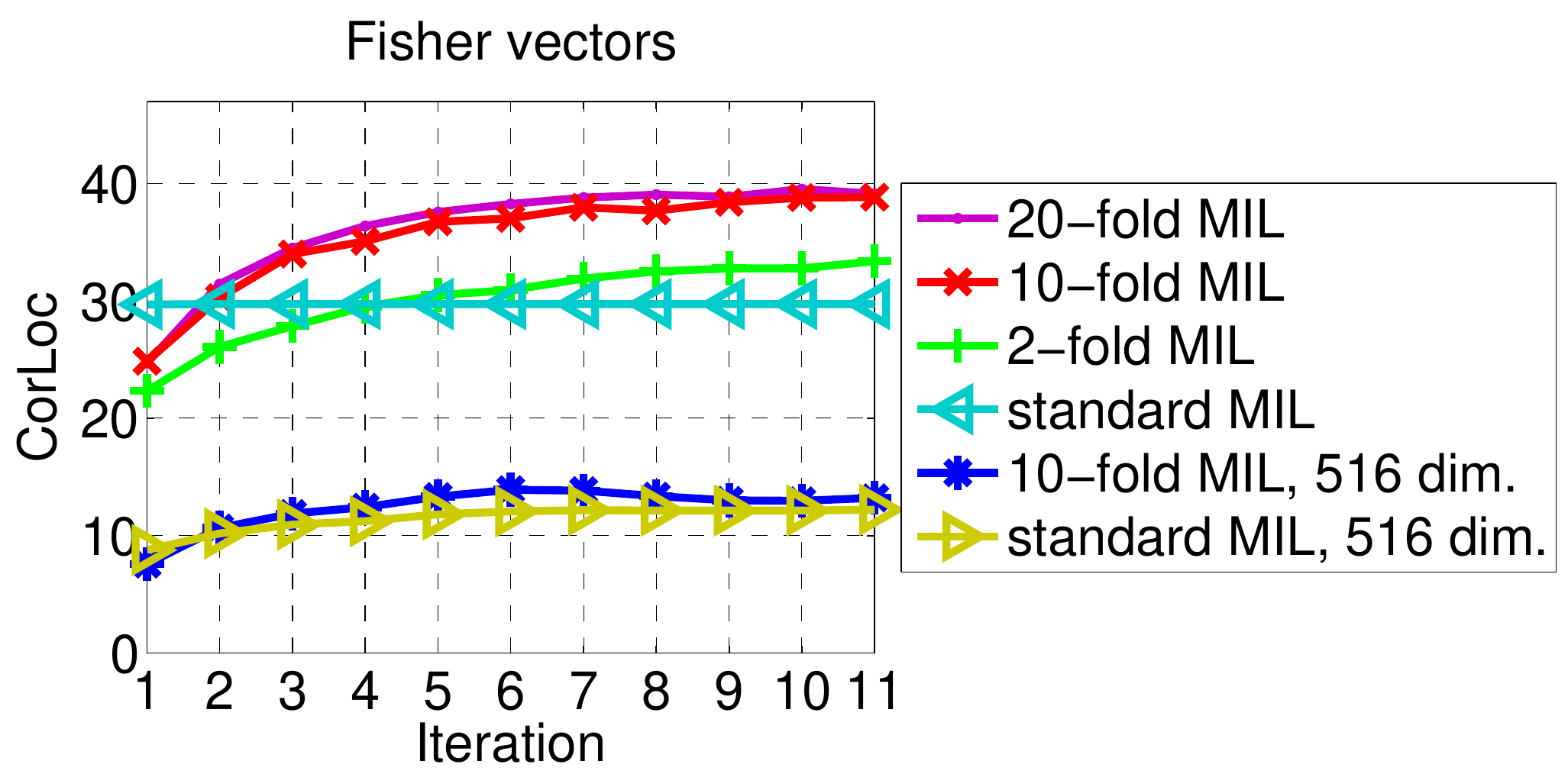} 
\includegraphics[height=.18\linewidth]{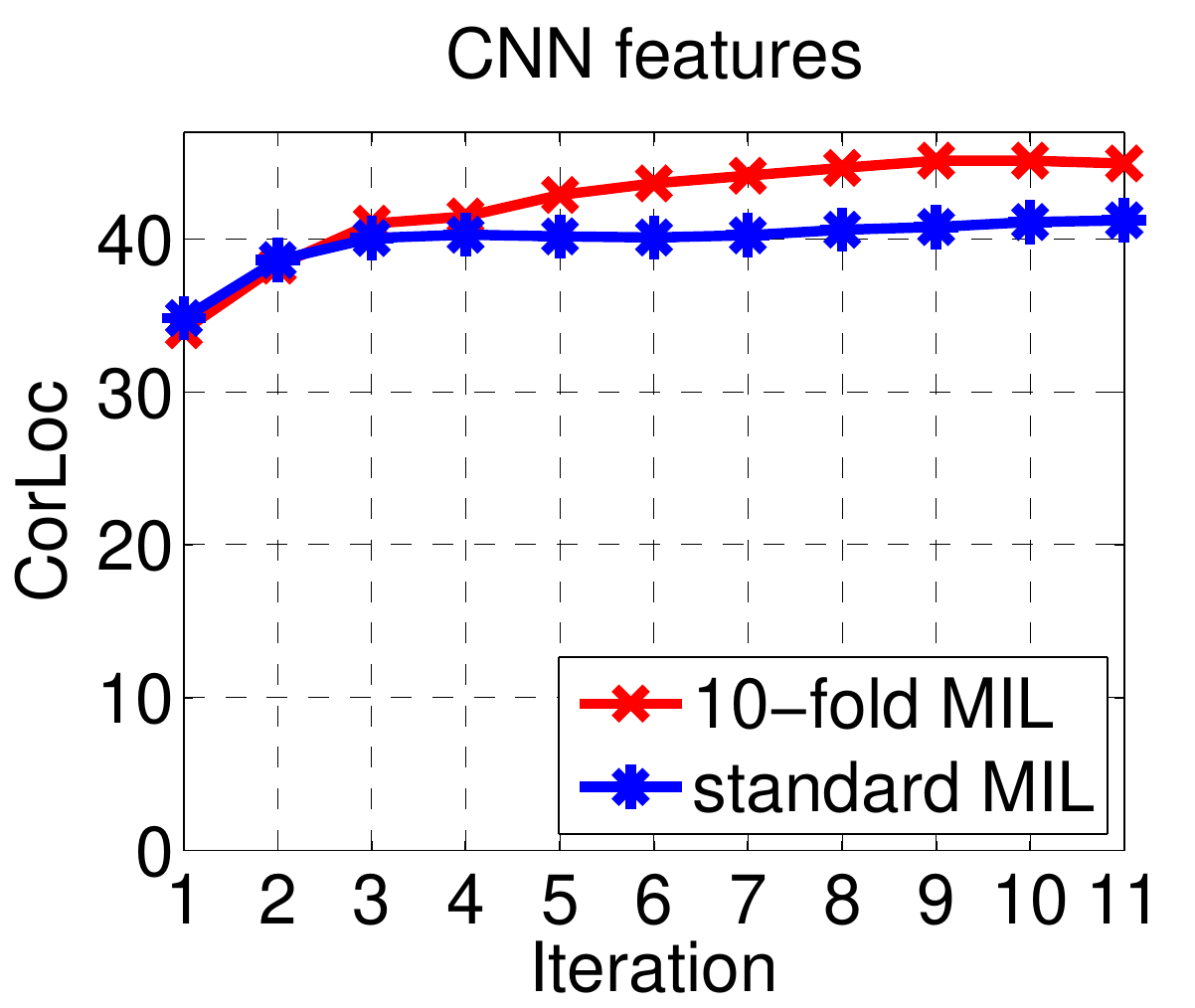} 
\includegraphics[height=.18\linewidth]{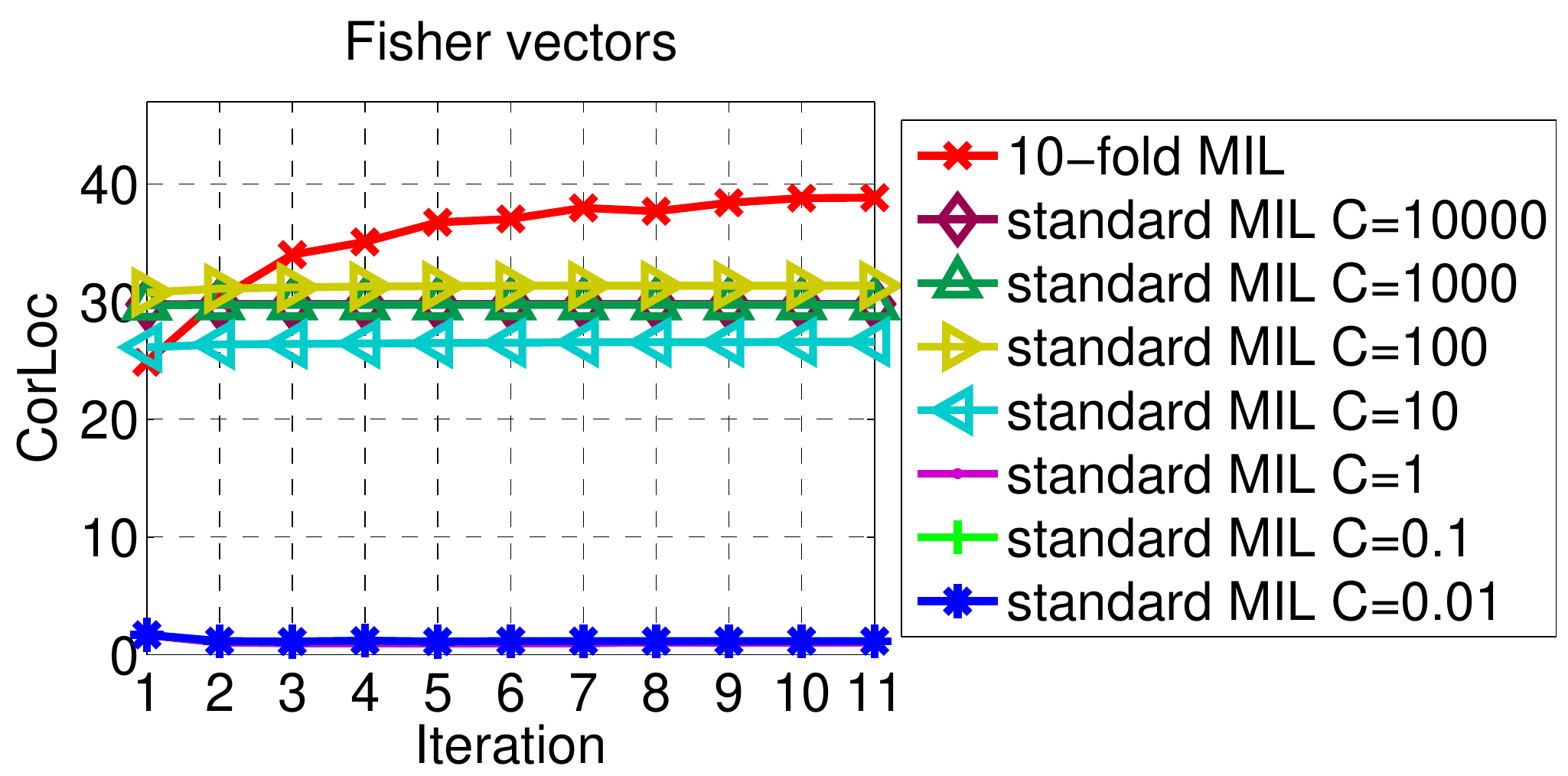} 
\end{center}
\caption{Correct localization (CorLoc) performance (in the training set) over the MIL
iterations, averaged across VOC 2007 classes. 
We show results for the high and low dimensional FVs (left panel), and the CNN features (middle panel). 
In the right panel, we compare 10-fold training  with standard MIL training using  different
values of the SVM cost parameter (C) for the high-dimensional FVs.}
\label{fig:avgcorlocs} 
\end{figure*}

The refined windows are likely to be better aligned with
object instances. Thus, their horizontal mirrors
are more reliable and can be used as additional training examples.
We evaluate the impact of window refinement and flipped
examples in the next section.

\section{Experimental evaluation}
\label{sec:experiments}

In this section we present a detailed analysis and
evaluation of our weakly supervised localization approach.

\subsection{Dataset and evaluation criteria}

We use the \VOC 2007 and 2010 datasets
\cite{everingham10ijcv} in our experiments.  Most of our
experiments use the 2007 dataset, which allows us to
compare to previous work. To the best of our knowledge, we
are the first to report WSL performance on the VOC 2010
dataset.  Following
\cite{deselaers12ijcv,pandey11iccv,shi13iccv}, during
training we discard any images that only contain object
instances marked as ``difficult'' or ``truncated''.
During testing all images are included.  We use linear SVM
classifiers, and set the weight of the regularization term
and the class weighting to fixed values based on
preliminary experiments. We perform two hard-negative
mining  steps~\cite{felzenszalb10pami} after each re-localization phase.
Finally, while we run all experiments using the same random seed,
we have empirically verified
that changing the seed does not affect the final
detection performance significantly.

Following \cite{deselaers12ijcv}, we assess performance
using two measures.  First, we evaluate the  fraction of
positive \emph{training images} in which we obtain correct
localization (CorLoc).  Second, we measure the
final object detection performance on the \emph{test images}
using the standard protocol~\cite{everingham10ijcv}, average precision (AP)
per class and mean AP (mAP) across all
classes.  For both measures, we consider that a window is
correct if it has an intersection-over-union ratio of at
least 50\% with a ground-truth object. 
Since CorLoc is not consistently measured across studies due to changes 
in training sets, we use CorLoc mainly as a diagnostic measure,
and use AP to compare to the state-of-the-art.

\subsection{Multi-fold MIL training and features}
\label{sec:DevEval}

In our first experiment, we compare (a) standard MIL
training, and (b) our multi-fold MIL algorithm with $K=10$
folds. Both are initialized from the full image up to the
4\% boundary. We also consider the effectiveness of
background features for the FV representation. We test
three variants: (F) foreground only descriptor, (B)
an FV computed over the window background, and (C)
our contrastive background descriptor. Finally, we
compare the FV representation to the CNN representation and the FV+CNN combination (by means of concatenating the descriptors).
Together, this yields ten combinations of features and
training algorithms. \tab{voc07_eval_corloc} presents
results in terms of CorLoc on the training set,
and \tab{voc07_eval_ap} presents results in terms of
AP on the test set.

From the results we see that, averaged over the classes,  multi-fold MIL outperforms standard MIL for all five tested representations, and for both CorLoc and AP.  Furthermore, we see that
the CorLoc differences across different FV descriptors are
rather small when using standard MIL training. This is due
to the degenerate re-localization performance with
high-dimensional descriptors for standard MIL training as
discussed in \sect{training}; we will come back to this
point below.  For multi-fold training, the CNN
features give better results than FV for 12 and 13 classes
in terms of CorLoc and AP, respectively. They also benefit
significantly from our multi-fold training procedure,
although to a lesser extent than the FV. This is due to
the lower dimensionality of the CNN features compared to
the FV features.

While the CNN features give better
performance overall than FV, we observe that the FV+CNN feature
combination improves over the individual features in 13
of 20 classes in terms of both CorLoc and AP scores using 
multi-fold MIL. 
Importantly, we note that standard MIL over the combined
feature space performs poorly at 34.4\% CorLoc and 22.0\%
mAP compared to
47.3\% CorLoc and 27.4\% mAP for multi-fold MIL.

\fig{imagesiter:fv} presents examples of
re-localization using standard and multi-fold MIL
training. In all three cases,
we observe that standard MIL gets stuck with the windows
found by the first re-localization step. In contrast,
multi-fold MIL is able to progressively localize down to
smaller image regions. In the {\em bicycle}
and {\em motorbike} examples, multi-fold MIL
successfully localizes the object instances.  In the {\em
cat} example, on the other hand, while the window
localized by standard MIL is correct, multi-fold MIL
localizes  the cat face, which has below 50\%
overlap with the object bounding box.  
The failure example in \fig{imagesiter:fv} affirms the 
difficulty for weakly supervised localization that we have
pointed out in \sect{refinement}: the WSL labels only
indicate to learn a model for the most repeatable
structure in the positive training images. For the cat class, due to the highly
deformable body, the head can be argued to  be
the most distinctive and reliably detectable structure.
This is what multi-fold MIL learns, but it degrades its
CorLoc and AP scores. Parkhi \etal~\cite{parkhi11iccv} also observed this, and proposed to localize cats and dogs based on a head detector in a fully supervised detector setting.  
Our window refinement method, which we evaluate below, resolves this issue to some extent.

In our next experiment, we further investigate the
localization performances of the algorithms in terms of
CorLoc across the training iterations. In the left panel
of \fig{avgcorlocs} we show the results for standard
MIL, and our multi-fold MIL algorithm using 2, 10, and 20
folds. The results clearly show the degenerate
re-localization performance obtained with standard MIL
training, of which CorLoc stays (almost) constant in the
iterations following the first re-localization stage. Our
multi-fold MIL approach leads to substantially better
performance, and ten MIL iterations suffice for the
performance to stabilize. Results increase significantly
by using 2-fold and 10-fold training respectively.  The
gain by using 20 folds is limited, however, and therefore
we use 10 folds in the remaining experiments. 
We also include experiments with the 516 dimensional FV
obtained using a 4-component MoG model, to verify the
hypothesis of \sect{training}.  The latter conjectured
that the degenerate re-localization observed for standard
MIL training is due to the trivial separability obtained
for high-dimensional descriptors.  Indeed, the lowest two
curves in the left panel of \fig{avgcorlocs} show that for
this descriptor we obtain non-degenerate re-localization
using standard MIL similar to multi-fold MIL.  The
performance is poor, however, due to limited
representative power of the low-dimensional FVs.

\def\myfigA#1{\adjustimage{width=30mm,height=21.4mm,valign=m,margin=0mm .6mm}{final_localization_comparison/images/#1}}
\def\myfig#1#2#3{\rotatebox[origin=c]{90}{{#1}} & \myfigA{#2/#3_fv_standard.pdf} & \myfigA{#2/#3_cnn_standard.pdf} & \myfigA{#2/#3_fv_kfold.pdf} & \myfigA{#2/#3_cnn_kfold.pdf} \\ \hline}
\begin{figure*}
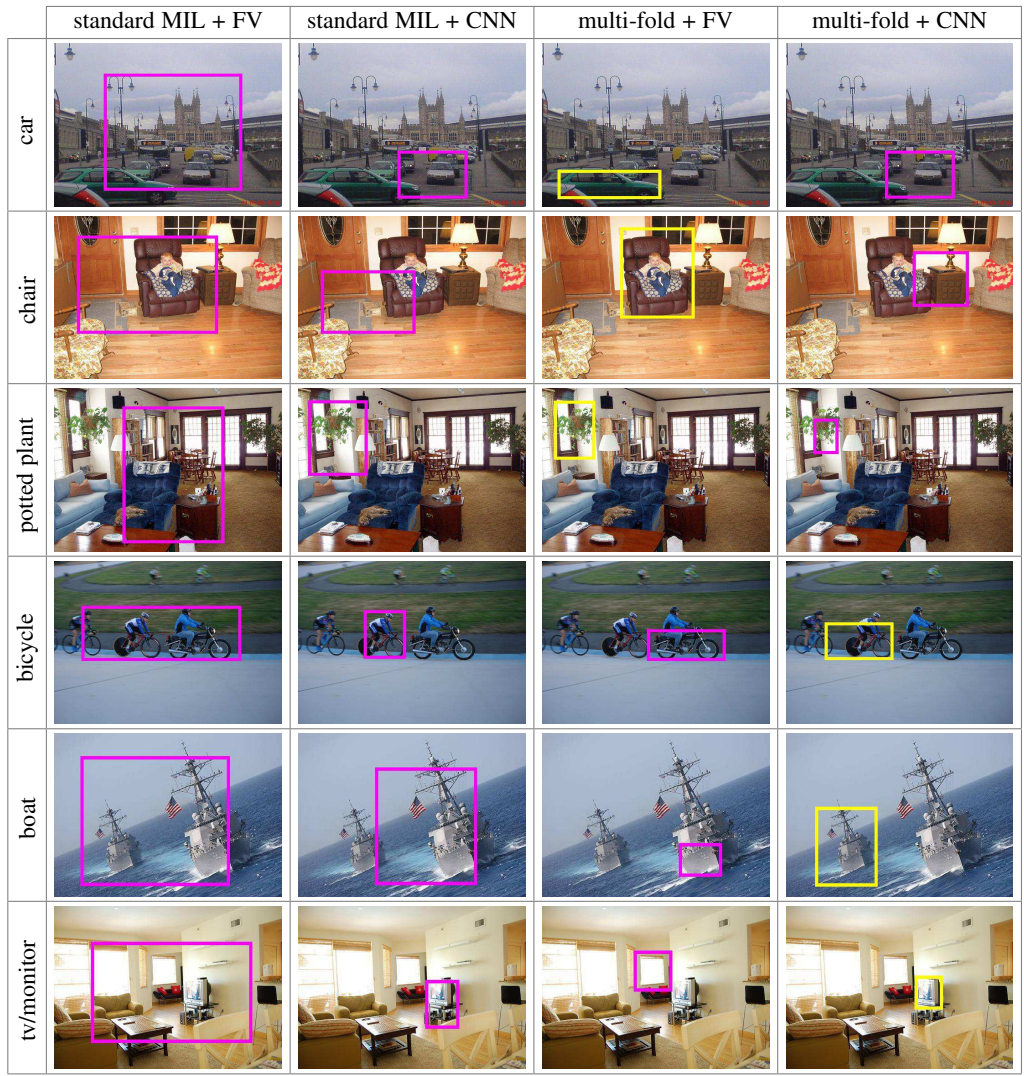

\centering
{\small\addtolength{\tabcolsep}{-3pt}
\begin{tabular}{|c|c|c|c|c|}
\arrayrulecolor[rgb]{.5,.5,.5}
\cline{2-5}
\multicolumn{1}{c|}{} & standard MIL + FV & standard MIL + CNN & multi-fold + FV & multi-fold + CNN \\ 
\hline
\myfig{car}{car}{003636} 
\myfig{chair}{chair}{005439} 
\myfig{potted plant}{pottedplant}{005794} 
\myfig{bicycle}{bicycle}{003855} 
\myfig{boat}{boat}{009347} 
\myfig{tv/monitor}{tvmonitor}{002989}
\end{tabular}
}
\caption{Example localization results on the training
images for standard MIL and
multi-fold MIL algorithms with high-dimensional FV and
CNN features. Correct localizations are shown in yellow, incorrect ones in pink.}
\label{fig:images:finalwnds}
\end{figure*}

In the middle panel of \fig{avgcorlocs}, we compare standard
MIL and multi-fold MIL using the CNN features.  We observe
that standard MIL is less affected by
degenerate re-localization problem, compared to the case
for high-dimensional FVs. This is in accordance with our
observations for low-dimensional FVs, as the CNN features
have an intermediate dimensionality of 4,096.
Nevertheless, multi-fold MIL leads to significant
improvements over the iterations, which results in 43.8\%
CorLoc, compared to 40.3\% CorLoc for standard MIL.

The degenerate re-localization of standard MIL using
high-dimensional descriptors can be interpreted as
over-fitting to the training data at an early stage.
Therefore, the question is whether we can improve standard
MIL by carefully tuning the trade-off between the
regularization terms and the loss functions for SVM
training. In the right panel of \fig{avgcorlocs}, we
investigate this question by evaluating the standard MIL
approach for different values of the cost parameter ($C$)
using high-dimensional FVs. The results show that,
although choosing a proper $C$ value is important, it is
not possible to solve the degenerate re-localization
problem of standard MIL in this manner.  Whereas using a
too low $C$ value ($C{\leq}1$) causes standard MIL to
drift off to a poor solution, larger $C$ values
($C{\geq}10$) result in degenerate re-localization.

\subsection{Evaluation of window refinement}

In \fig{images:finalwnds} we provide examples of the
localization results on the training images using standard
and multi-fold MIL for  FV and  CNN features.
The first three examples ({\em car}, {\em chair}, and {\em
potted plant}) are  only  correctly localized using
multi-fold MIL with FVs. 
These examples demonstrate the ability of our multi-fold training procedure to handle cases
with multiple instances that appear in near proximity and
with considerable background clutter. 
The last three examples ({\em bicycle}, {\em boat}, and
{\em tv/monitor}) are  only  correctly localized using multi-fold MIL with CNNs. 
  In the {\em bicycle}
example, we observe that multi-fold MIL with FVs mistakes
a visually similar motorbike for a bicycle.
Likewise, in the {\em tv/monitor} example, multi-fold MIL
over FVs localizes a window that looks similar to a bright
monitor. 
These examples  suggest that (i) FV and
CNN features can be complimentary to each other, which gives an insight for the success of the FV+CNN representation, and (ii) that some near-miss localizations might be corrected by a  window refinement method.

We present the CorLoc and AP results for the window refinement method in
\tab{voc07_refinement_corloc} and \tab{voc07_refinement_ap}, respectively.  
All reported results are based on multi-fold training.
The upper parts of the tables show the results for multi-fold MIL
without window refinement, therefore, contain copies of the corresponding rows from \tab{voc07_eval_corloc}  and \tab{voc07_eval_ap}. The bottom parts show the results for the window
refinement method for the FV, CNN and the combined features. We observe that the 
refinement method significantly improves the 
average CorLoc and AP scores for all three window descriptor types.
In the case of FV+CNN features, applying the window refinement method improves CorLoc and detection AP 
in 16 out of 20 classes, where we measure the largest three
improvements in CorLoc for the classes {\em horse}, {\em dog} and
{\em cat}. The instances of these three classes have
deformable shapes, therefore, the weakly supervised
localization tends to result in imprecise localizations or
part localizations, some of which are corrected by the
window refinement method. The four classes for which the
window refinement method deteriorates CorLoc  are {\em
bicycle,  bottle, chair} and {\em potted-plant}.  These
classes typically have highly textured and/or small
instances, where the edge-driven objectness measure can be
misleading. Finally, we note that the results obtained with refinement also
include the addition of horizontal flips of the positive
training windows. This has only a minor effect on
performance: without these the detection mAP for the FV+CNN features drops by only
0.4\% to 29.8\%. Overall, these results show that the FV and CNN
features are complimentary, and that window refinement can
improve localization performance.

\subsection{Comparison to state-of-the-art WSL detection}
\label{sec:CompSota}

We compare  our multi-fold MIL approach
to the state-of-the-art in terms of detection AP in \tab{stateofart_ap}.  We separate
the recent work into two groups in terms of their
utilization of auxiliary training data.  
To the best our knowledge, only three previous studies  that do not
use auxiliary training data reported 
detection AP scores on PASCAL VOC 2007. Other work, such as \eg that of Deselaers \etal~\cite{deselaers12ijcv}, was evaluated only under
simplified conditions, such as using viewpoint information
and using images from a limited number of classes.
Russakovsky \etal \cite{russakovsky12eccv} report mAP over
all 20 classes, but report separate AP values for only six
classes.  Multi-fold MIL over the FV-only features with window refinement, results
in a detection mAP of 23.3\%, which is significantly
better than the 13.9\% and 15.0\% reported in ~\cite{siva11iccv} and \cite{russakovsky12eccv}.

The second half of \tab{stateofart_ap} presents the recent
work that uses CNN-based features, which involves
representation learning on the ImageNet dataset. For
comparison, we use our multi-fold MIL approach over the
FV+CNN features with window refinement.  Our detection mAP
of 30.2\% is significantly better than the 22.7\%  and 24.6\% by Song \etal
\cite{song14icml,song14nips}, and the 26.4\% by Bilen \etal
\cite{bilen14bmvc}.  Wang \etal~\cite{wang14eccv} report a
detection mAP of 30.9\%, and additionally an improved mAP
of 31.6\% mAP using the contextual rescoring method of
\cite{felzenszalb10pami}.  Our detection mAP is comparable
to Wang \etal~\cite{wang14eccv} without inter-class
context, and we obtain better AP scores in 11 out of 20
classes. 

{\addtolength{\tabcolsep}{-3.5pt}
\begin{table*}
\caption{Evaluation of window refinement on the VOC 2007 dataset, in terms of training set localization accuracy (CorLoc).}
\label{tab:voc07_refinement_corloc}
\begin{center}
\begin{tabular}{|l|cccccccccccccccccccc|c|}
\cline{2-22}
\multicolumn{1}{l|}{} & aero            & bicy            & bird                & boa                 & bot                 & bus             & car             & cat             & cha                 & cow             & dtab            & dog                 & hors                & mbik            & pers            & plnt            & she                 & sofa            & trai            & tv              & Av.\tabularnewline
\hline
FV                    & {56.6}          & 58.3            & 28.4                & 20.7                & 6.8                 & 54.9            & 69.1            & 20.8            & 9.2                 & 50.5            & 10.2            & 29.0                & {58.0}              & 64.9            & {{36.7}}        & 18.7            & 56.5                & 13.2            & 54.9            & {{59.4}}        & 38.8 \tabularnewline
\hline
CNN                   & 53.2            & {\textbf{66.7}} & 51.3                & 31.7                & 19.3                & {\textbf{70.5}} & 72.0            & 23.3            & 24.9                & 62.1            & 32.7            & 28.0                & 54.6                & 64.9            & 22.1            & 39.0            & 55.1                & 33.0            & 54.9            & 40.1            & 45.0 \\
\hline
FV+CNN                & 57.2            & 62.2            & 50.9                & 37.9                & {{{\textbf{23.9}}}} & 64.8            & 74.4            & 24.8            & {{{\textbf{29.7}}}} & 64.1            & 40.8            & 37.3                & 55.6                & 68.1            & 25.5            & {{38.5}}        & 65.2                & 35.8            & 56.6            & 33.5            & 47.3 \\
\hline
 & \multicolumn{20}{c}{after window refinement} & \tabularnewline
\hline
FV                    & 62.4            & 62.2            & 40.7                & 35.2                & 5.1                 & 67.2            & 76.9            & 33.2            & 12.9                & 63.1            & 16.3            & 39.4                & 62.8                & 67.6            & 37.2            & 22.5            & 63.8                & 22.6            & {\textbf{65.5}} & {\textbf{65.5}} & 46.1 \\
\hline
CNN                   & {\textbf{67.1}} & 66.1            & 49.8                & 34.5                & 23.3                & 68.9            & {\textbf{83.5}} & {\textbf{44.1}} & 27.7                & {\textbf{71.8}} & {\textbf{49.0}} & 48.0                & 65.2                & {\textbf{79.3}} & {\textbf{37.4}} & {\textbf{42.9}} & 65.2                & {\textbf{51.9}} & 62.8            & 46.2            & {\textbf{54.2}} \\
\hline
FV+CNN                & {{65.3}}        & 55.0            & {{{\textbf{52.4}}}} & {{{\textbf{48.3}}}} & 18.2                & 66.4            & {{77.8}}        & {{35.6}}        & 26.5                & {{67.0}}        & {{46.9}}        & {{{\textbf{48.4}}}} & {{{\textbf{70.5}}}} & {{69.1}}        & {35.2}          & 35.2            & {{{\textbf{69.6}}}} & {{43.4}}        & {{64.6}}        & {43.7}          & {{52.0}} \\
\hline
\end{tabular}
\end{center}
\end{table*}
}

{\addtolength{\tabcolsep}{-3.5pt}
\begin{table*}
\caption{Evaluation of window refinement on the VOC 2007 dataset, in terms of test-set average precision (AP).}
\label{tab:voc07_refinement_ap}
\begin{center}
\begin{tabular}{|l|cccccccccccccccccccc|c|}
\cline{2-22}
\multicolumn{1}{l|}{} & aero            & bicy          & bird            & boa                   & bot                   & bus                   & car                   & cat             & cha                   & cow             & dtab                  & dog                   & hors                  & mbik                  & pers            & plnt                  & she                   & sofa                  & trai                  & tv                & Av.\tabularnewline
\hline
FV                    & 35.8            & 40.6          & 8.1             & 7.6                   & 3.1                   & 35.9                  & 41.8                  & 16.8            & 1.4                   & 23.0            & 4.9                   & 14.1                  & 31.9                  & {{41.9}}              & {{19.3}}        & 11.1                  & 27.6                  & 12.1                  & 31.0                  & {{\textbf{40.6}}} & 22.4 \tabularnewline
\hline
CNN                   & 32.1            & {46.9}        & {28.4}          & {12.0}                & {9.6}                 & 39.4                  & 45.5                  & 16.2            & {14.8}                & {33.1}          & {11.6}                & 14.0                  & 31.2                  & 39.3                  & 13.1            & {19.7}                & {30.5}                & {23.4}                & 37.0                  & 19.6              & {25.9} \\
\hline
FV+CNN                & 38.1            & \textbf{47.6} & 28.2            & 13.9                  & {{{{\textbf{13.2}}}}} & 45.2                  & {{{{\textbf{48.0}}}}} & 19.3            & {{{{\textbf{17.1}}}}} & 27.7            & 17.3                  & 19.0                  & 30.1                  & 45.4                  & 13.5            & 17.0                  & 28.8                  & 24.8                  & 38.2                  & 15.0              & 27.4 \\
\hline
 & \multicolumn{20}{c}{after window refinement} & \tabularnewline
\hline
FV                    & 36.9            & 38.3          & 11.5            & 11.1                  & 1.0                   & 39.8                  & 45.7                  & 16.5            & 1.2                   & 26.4            & 4.3                   & 17.7                  & 31.8                  & 44.0                  & 13.1            & 11.0                  & 31.4                  & 9.7                   & 38.5                  & 36.9              & 23.3 \\
\hline
CNN                   & {\textbf{40.4}} & 43.5          & {\textbf{29.5}} & 11.4                  & 9.4                   & 42.2                  & 47.3                  & {\textbf{25.6}} & 7.6                   & {\textbf{33.8}} & 15.8                  & 27.7                  & 37.4                  & 46.4                  & {\textbf{20.5}} & 19.9                  & 30.2                  & 23.5                  & 40.6                  & 19.6              & 28.6 \\
\hline
FV+CNN                & {{{39.3}}}      & 43.0          & {{{28.8}}}      & {{{{\textbf{20.4}}}}} & 8.0                   & {{{{\textbf{45.5}}}}} & 47.9                  & {{22.1}}        & 8.4                   & {{{33.5}}}      & {{{{\textbf{23.6}}}}} & {{{{\textbf{29.2}}}}} & {{{{\textbf{38.5}}}}} & {{{{\textbf{47.9}}}}} & {{{20.3}}}      & {{{{\textbf{20.0}}}}} & {{{{\textbf{35.8}}}}} & {{{{\textbf{30.8}}}}} & {{{{\textbf{41.0}}}}} & 20.1              & {{\textbf{30.2}}} \\
\hline
\end{tabular}
\end{center}
\end{table*}
}

{\addtolength{\tabcolsep}{-3.5pt}
\begin{table*}
\begin{center}
\begin{tabular}{|l|cccccccccccccccccccc|c|}
\cline{2-22}
\multicolumn{1}{l|}{}                          & aero                & bicy            & bird              & boa                 & bot                 & bus             & car               & cat                 & cha                   & cow                   & dtab              & dog                   & hors              & mbik                  & pers              & plnt                  & she                   & sofa                  & trai            & tv                  & Av.\tabularnewline
\hline
Pandey and Lazebnik'11  \cite{pandey11iccv}    & 11.5                & ---             & ---               & 3.0                 & ---                 & ---             & ---               & ---                 & ---                   & ---                   & ---               & ---                   & 20.3              & 9.1                   & ---               & ---                   & ---                   & ---                   & 13.2            & ---                 & ---\tabularnewline
\hline
Siva and Xiang'11  \cite{siva11iccv}           & 13.4                & 44.0            & 3.1               & 3.1                 & 0.0                 & 31.2            & 43.9              & 7.1                 & 0.1                   & 9.3                   & 9.9               & 1.5                   & 29.4              & 38.3                  & 4.6               & 0.1                   & 0.4                   & 3.8                   & 34.2            & 0.0                 & 13.9 \tabularnewline
\hline
Russakovsky \etal'12  \cite{russakovsky12eccv} & 30.8                & 25.0            & ---               & 3.6                 & ---                 & 26.0            & ---               & ---                 & ---                   & ---                   & ---               & ---                   & 21.3              & 29.9                  & ---               & ---                   & ---                   & ---                   & ---             & ---                 & 15.0 \tabularnewline
\hline
Ours (FV-only)                                 & 36.9                & 38.3            & 11.5              & 11.1                & 1.0                 & 39.8            & 45.7              & 16.5                & 1.2                   & 26.4                  & 4.3               & 17.7                  & 31.8              & 44.0                  & 13.1              & 11.0                  & 31.4                  & 9.7                   & 38.5            & 36.9                & 23.3 \\
\hline
  & \multicolumn{20}{c}{methods using additional training data} & \tabularnewline
\hline
Song \etal'14  \cite{song14icml}               & 27.6                & 41.9            & 19.7              & 9.1                 & 10.4                & 35.8            & 39.1              & 33.6                & 0.6                   & 20.9                  & 10.0              & 27.7                  & 29.4              & 39.2                  & 9.1               & 19.3                  & 20.5                  & 17.1                  & 35.6            & 7.1                 & 22.7 \tabularnewline
\hline
Song \etal'14 \cite{song14nips}                & 36.3                & {\textbf{47.6}} & 23.3              & 12.3                & 11.1                & 36.0            & 46.6              & 25.4                & 0.7                   & 23.5                  & 12.5              & 23.5                  & 27.9              & 40.9                  & 14.8              & 19.2                  & 24.2                  & 17.1                  & 37.7            & 11.6                & 24.6 \\
\hline
Bilen \etal'14 \cite{bilen14bmvc}              & 42.2                & {43.9}          & 23.1              & 9.2                 & {{{\textbf{12.5}}}} & {44.9}          & 45.1              & 24.9                & 8.3                   & 24.0                  & 13.9              & 18.6                  & 31.6              & 43.6                  & 7.6               & {{{{\textbf{20.9}}}}} & 26.6                  & 20.6                  & 35.9            & 29.6                & 26.4 \tabularnewline
\hline
Wang \etal'14 \cite{wang14eccv}                & 48.8                & 41.0            & 23.6              & {12.1}              & 11.1                & 42.7            & 40.9              & {{{\textbf{35.5}}}} & {{{{\textbf{11.1}}}}} & {{{{\textbf{36.6}}}}} & 18.4              & {{{{\textbf{35.3}}}}} & 34.8              & 51.3                  & 17.2              & 17.4                  & 26.8                  & 32.8                  & 35.1            & 45.6                & 30.9 \tabularnewline
\hline
Wang \etal'14 \cite{wang14eccv} +context       & {{{\textbf{48.9}}}} & 42.3            & {26.1}            & 11.3                & 11.9                & 41.3            & 40.9              & 34.7                & 10.8                  & 34.7                  & {{18.8}}          & 34.4                  & {{35.4}}          & {{{{\textbf{52.7}}}}} & {{19.1}}          & 17.4                  & {{{{\textbf{35.9}}}}} & {{{{\textbf{33.3}}}}} & 34.8            & {{{\textbf{46.5}}}} & {{{\textbf{31.6}}}} \tabularnewline
\hline
Ours                                           & 39.3                & 43.0            & {{\textbf{28.8}}} & {{{\textbf{20.4}}}} & 8.0                 & {\textbf{45.5}} & {{\textbf{47.9}}} & 22.1                & 8.4                   & {33.5}                & {{\textbf{23.6}}} & 29.2                  & {{\textbf{38.5}}} & {47.9}                & {{\textbf{20.3}}} & {20.0}                & {35.8}                & {30.8}                & {\textbf{41.0}} & 20.1                & {30.2} \\
\hline
\end{tabular}
\end{center}
\caption{Comparison of WSL  detectors
on PASCAL VOC 2007 in terms of test-set detection AP.  
Results for Pandey and Lazebnik \cite{pandey11iccv} are
taken from \cite{prest12cvpr}. 
}
\label{tab:stateofart_ap}
\end{table*}
}

\subsection{Analysis of performance and failure cases}

{\addtolength{\tabcolsep}{-4pt}
\begin{table}
\caption{Performance in test-set detection mAP on VOC 2007 using FV, CNN and FV+CNN features, with varying degrees of supervision.}
\centering
\resizebox{\linewidth}{!}{
{ \footnotesize
\begin{tabular}{|c|c|c|c|c|c|}
\hline
Supervision               & Neg on Pos & Positive Set   & FV & CNN & FV+CNN \tabularnewline
\hline 
\hline 
Image labels only      & No              & Non-diff/trunc & 22.4 & 25.9 & 27.4 \tabularnewline
\hline
Cand box for one obj  & No              & Non-diff/trunc & 30.8 & 36.5 & 40.5 \tabularnewline 
\hline
Cand box for all obj  & No              & Non-diff/trunc & 30.7 & 35.7 & 38.4 \tabularnewline
\hline
Cand box for all obj  & Yes             & Non-diff/trunc & 32.0 & 41.2 & 43.7 \tabularnewline 
\hline
Exact box for all obj & Yes             & Non-diff/trunc & 32.8 & 40.5 & 43.6 \tabularnewline  
\hline
Exact box for all obj & Yes             & All            & 35.4 & 42.8 & 46.2 \tabularnewline  
\hline
\end{tabular}
}
}
\label{tab:spectrum}
\end{table}
}

To analyze the causes of difficulty of WSL for object
detection, we consider the performance of our detector
when used in a fully supervised training setting.  For the
sake of brevity, we analyze the WSL results without applying 
window refinement.

There are several factors that change between the WSL and
fully supervised training. In order to evaluate the
importance of each factor, we progressively move from
the original WSL setting to the fully supervised setting.
In \tab{spectrum}, we report the resulting mAP values for
each step using the FV-only, CNN-only and FV+CNN features
in the final three columns, respectively.

In WSL we have to determine the object locations in the
positive training images. If in each positive training
image we fix the object hypothesis to the candidate window
that best overlaps with one of the ground-truth objects,
we no longer need to use MIL training. In this case, we
increase the detection mAP by 13.1 points to 40.5 \wrt the weakly supervised setting; see first and second row of 
\tab{spectrum}. Even though this is a significant
improvement \wrt WSL, there is still a gap of 5.7\% in detection mAP compared to the fully supervised setting. 

The remaining difference in performance is due to several
factors, we list them now and give the performance
improvements when making the WSL training scenario
progressively more similar to the supervised one.  (i) In WSL
only one instance per positive training image is used.
Including all instances instead makes a relatively minor effect on
the performance, see the third row in \tab{spectrum}. (ii)~In WSL
hard-negative mining is based on negative images 
only, when positive images are used too performance rises
to 43.7\% mAP for the FV+CNN features,
as shown in the fourth row. (iii) WSL is
based on the candidate windows, using the ground-truth
windows instead makes a relatively small impact, see the fifth row. (iv) Finally, in WSL, we do not use
positive training images marked as difficult or truncated,
if these are added 
performance rises to 46.2\% mAP for FV+CNN features.
\footnote{Note that our CNN-only fully-supervised mAP of
42.8\% is comparable
to the 44.7\% of Girshick \etal~\cite{girshick14cvpr}, which
uses similar CNN features.}

These results show that the most important two factors
are the use 
of correct training windows and hard-negative mining on
positive training images.  We also observe that 
multi-fold MIL achieves 59\% of the representational
performance limit (27.4\% out of 46.2\% mAP).
With respect to the 40.5\% mAP 
for training from ideal localizations, multi-fold MIL
approach attains 68\% of the WSL performance limit. Standard MIL (22.0\% mAP, \cf \tab{voc07_eval_ap})
attains only 54\% of this performance limit.

\begin{figure}
\def\myfig#1{\includegraphics[height=0.37\linewidth]{plotAP/Version3_Journal/#1}}
\begin{center}
\myfig{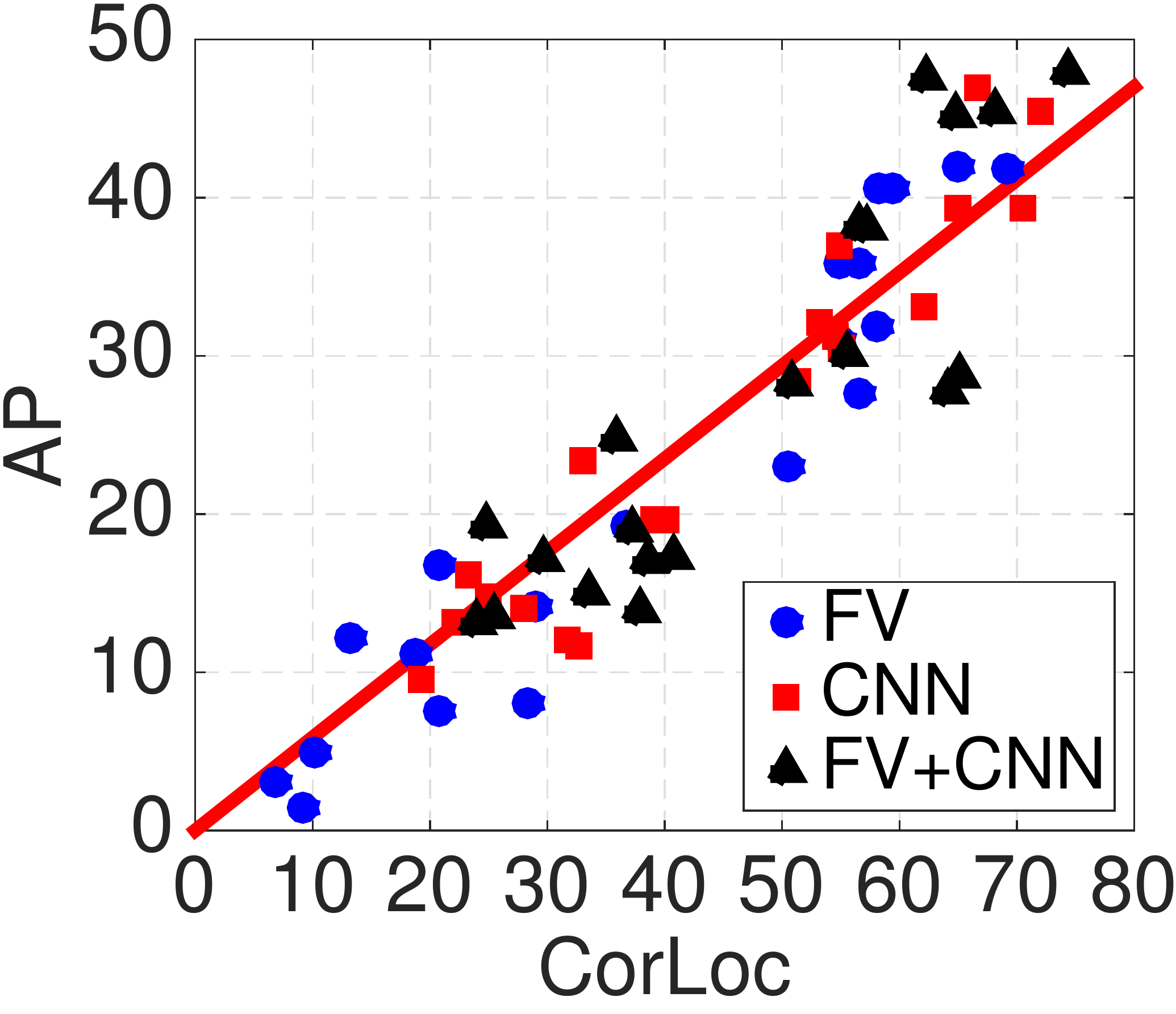} 
\myfig{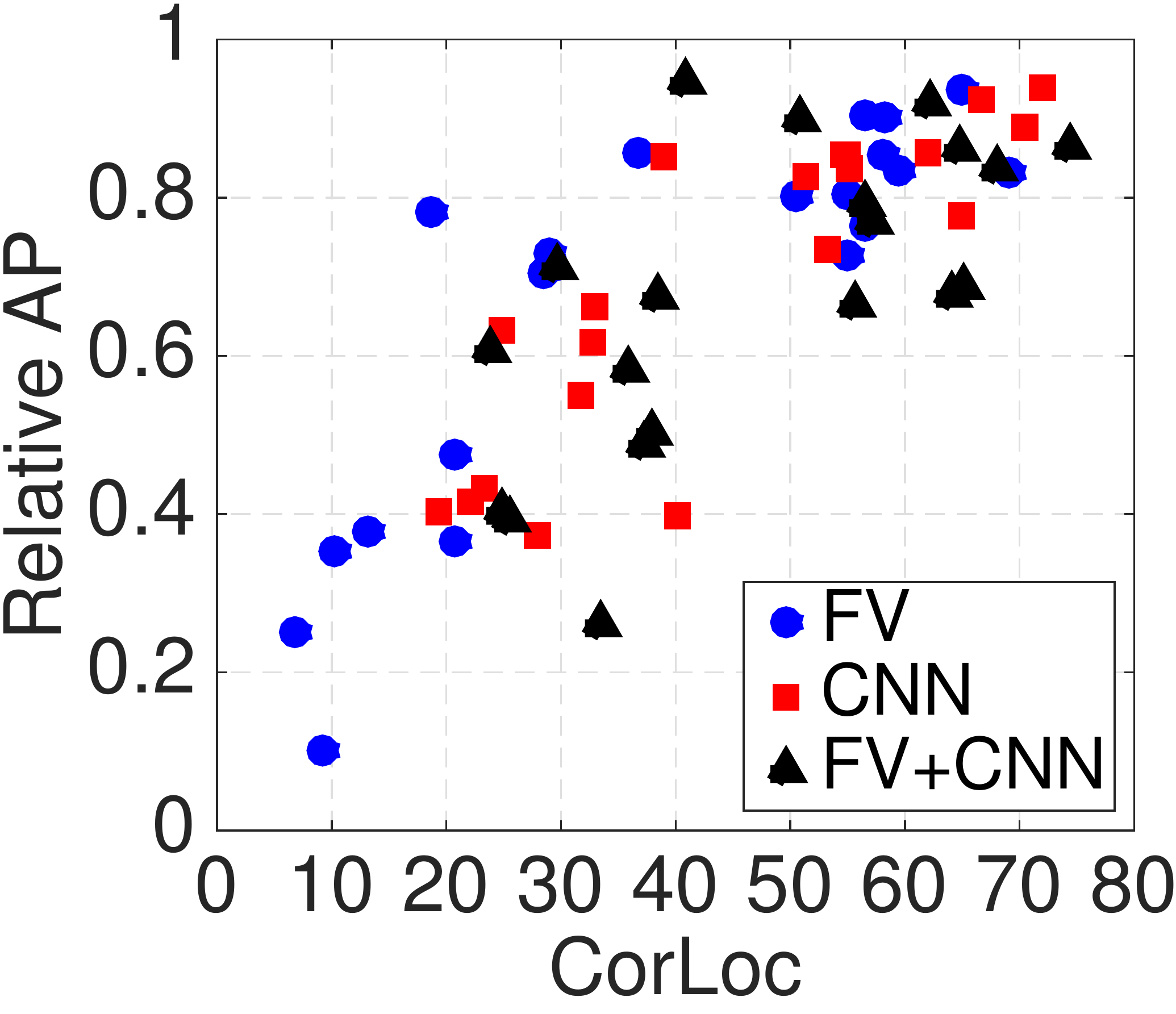}
\end{center}
\caption{AP \vs CorLoc for multi-fold MIL (left), and
ratio of WSL over supervised AP as a function of CorLoc
(right) using the FV (blue circles), CNN (red squares) and FV+CNN (black triangles) representations.
CorLoc and AP are measured on the training and test images, respectively.
The left plot shows the line with least squares error for the data points.}
\label{fig:CorLocAP}
\end{figure}

\def\myfig#1{\includegraphics[width=0.37\linewidth]{breakdown2_fvcnn/#1}} 

\begin{figure*}
\centering
{\centering
    \includegraphics[width=0.57\linewidth]{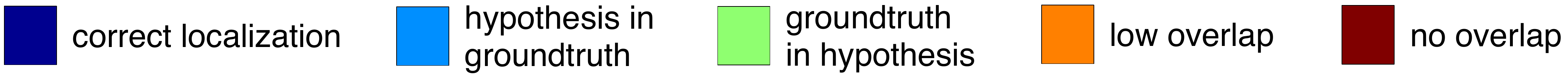} \\ \vspace{-1mm} 
    \subfloat[multi-fold MIL]{
        {\centering
            \myfig{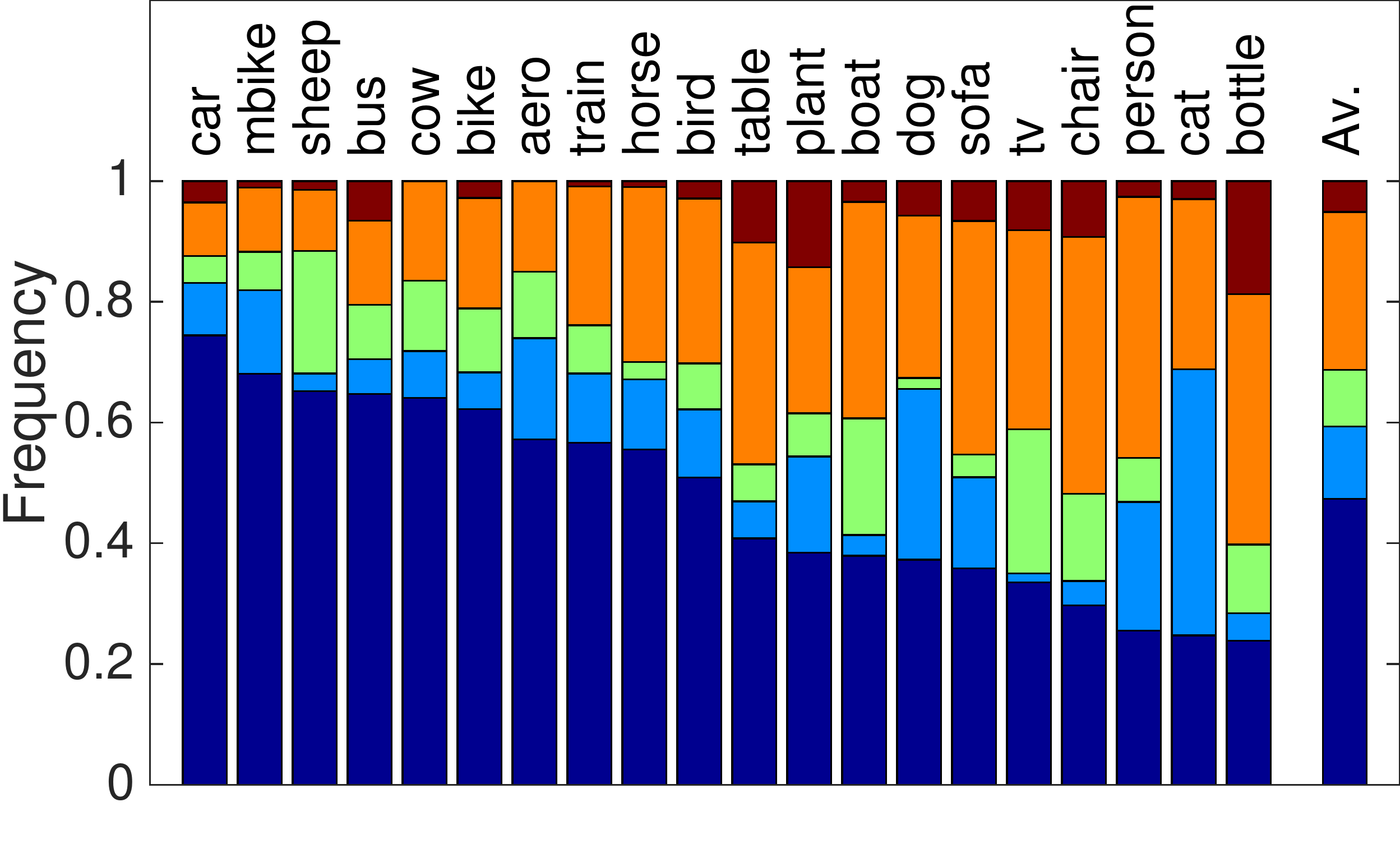}  
        }
        \label{fig:breakdown_all_multifold}
    }
    \subfloat[standard MIL]{
        {\centering
            \myfig{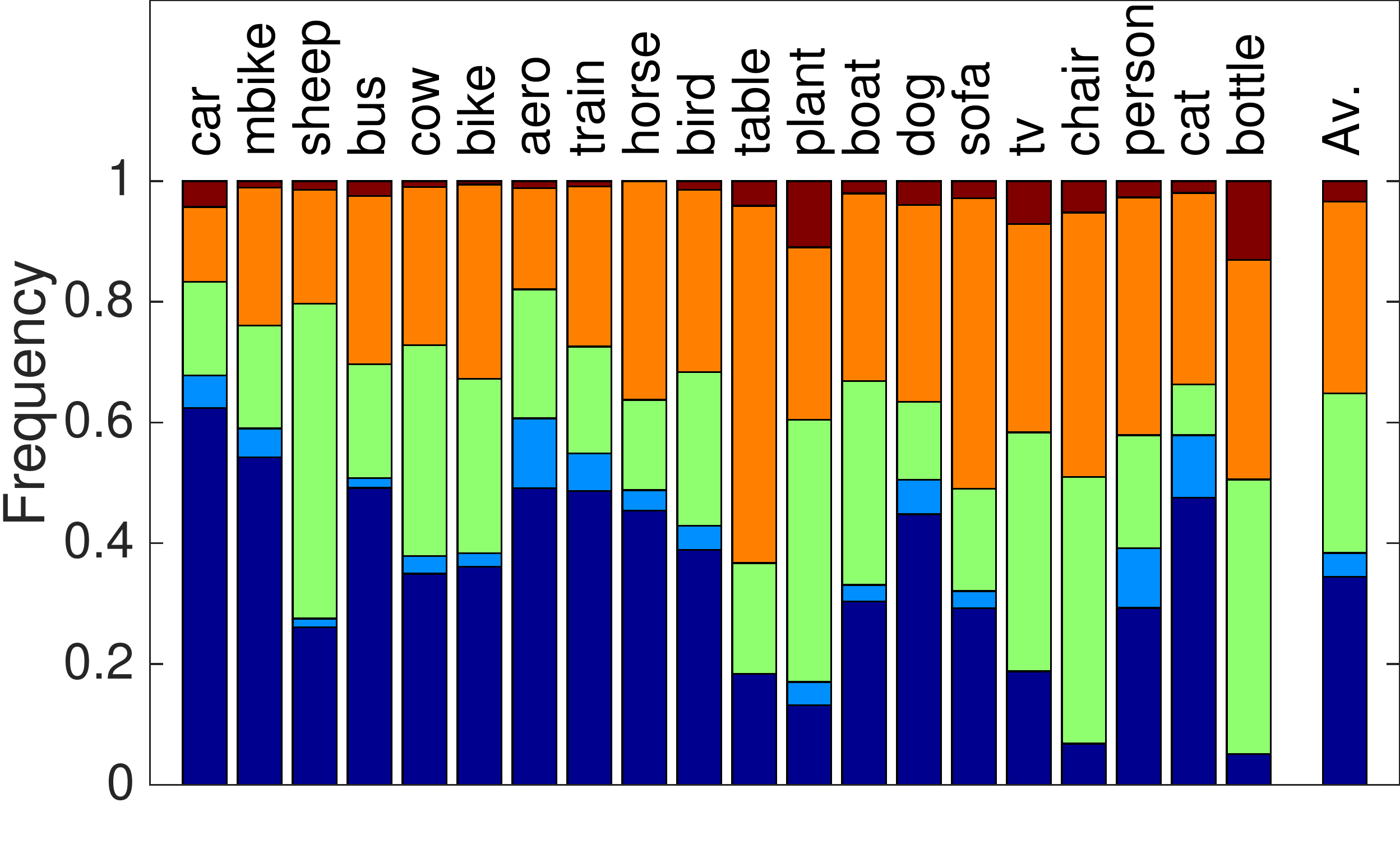} 
        }
        \label{fig:breakdown_all_standard}
    }
}
\caption{Per-class frequency of error modes, and averaged across all classes using FV+CNN features with 10-fold MIL and standard MIL training.}
\label{fig:breakdown_all}
\end{figure*}

In \fig{CorLocAP} we further analyze the results of our
weakly supervised detector, and its relation to the
optimally localized version. In the left panel, we
visualize the close relationship between the per-class
CorLoc and AP values for our multi-fold MIL detector.  The
three classes with lowest \mbox{CorLoc}  are \emph{bottle, chair}, and \emph{dining
table} using FVs, \emph{bottle, chair}, and \emph{cat} using CNNs,
and \emph{bottle, cat}, and \emph{person} using the FV+CNN combination.
Most instances of these classes appear in highly cluttered indoor images,
and are often occluded by objects (\emph{dining table, chair}),
or have extremely variable
appearance due to transparency (\emph{bottle}) and deformation (\emph{cat, person}).
In the right panel, we plot the ratio between our WSL detection AP
and the AP obtained with the same
detector trained with optimal localization
(the second row in \tab{spectrum}).
In this case there is also a clear relation
with our CorLoc values. 
The relation is quite different,
however, below and above 50\% \mbox{CorLoc}. Below this
threshold, due to the amount of noisy training examples,
WSL tends to break down. Above this
threshold, however, the training is able to cope with the
noisy positive training examples, and the
weakly supervised detector performs relatively well: on
average above 80\% relative to optimal localization.

\def\myfig#1{\includegraphics[width=0.19\linewidth]{mixedsupervised/journal_ratio/FV/percent_fig#1}}
\begin{figure*}
\begin{center}
\myfig{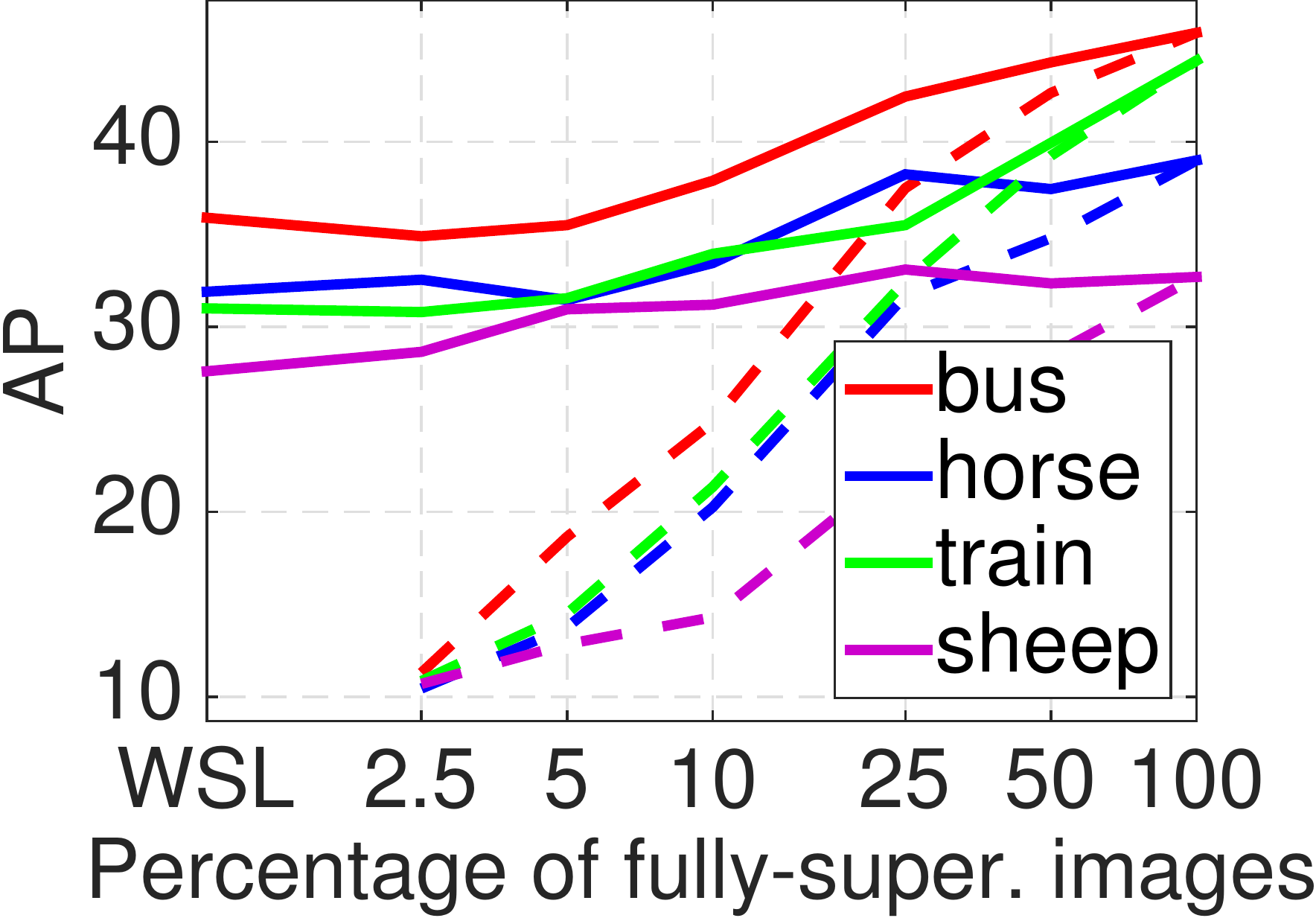} \myfig{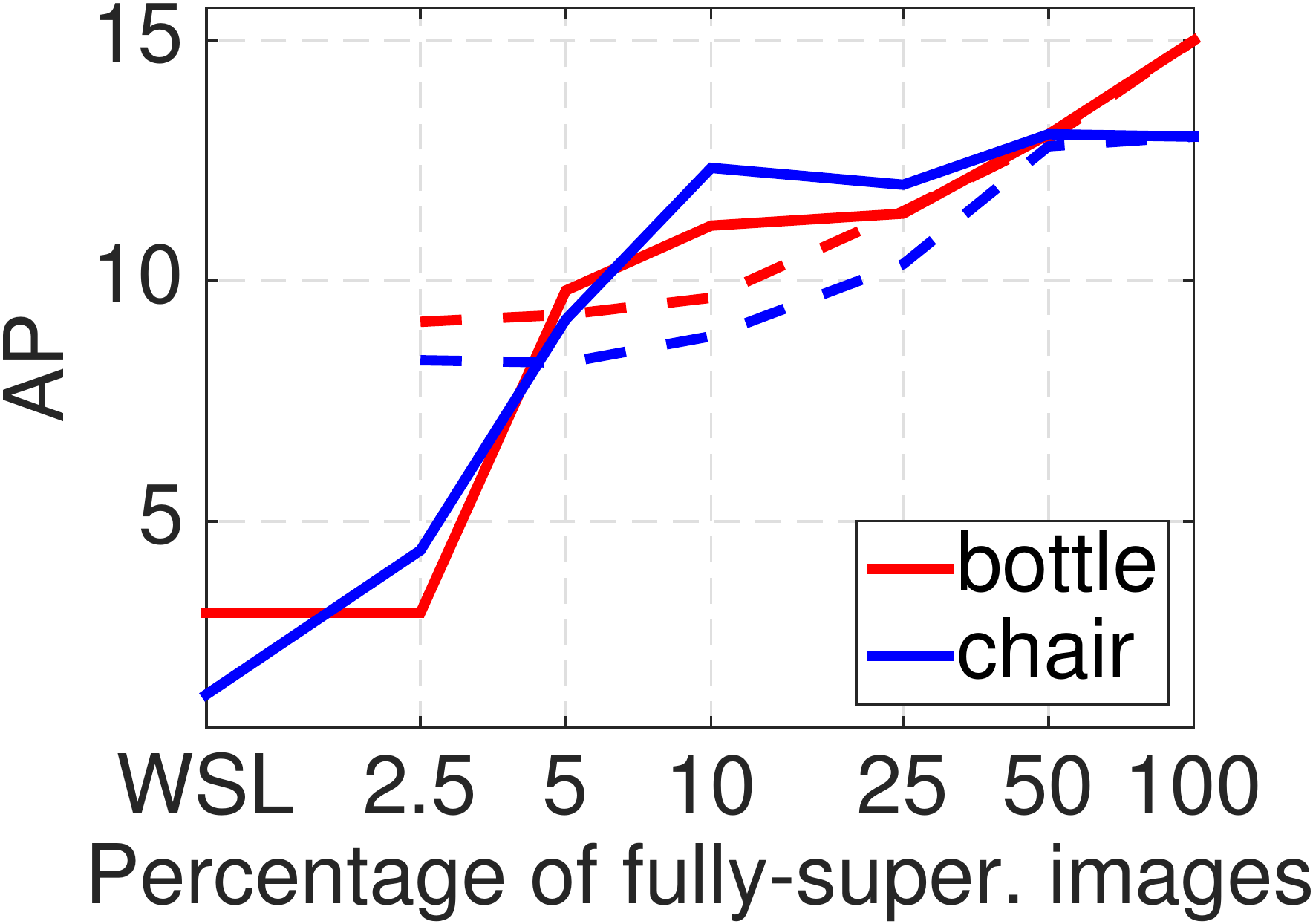} \myfig{9} \includegraphics[width=0.19\linewidth]{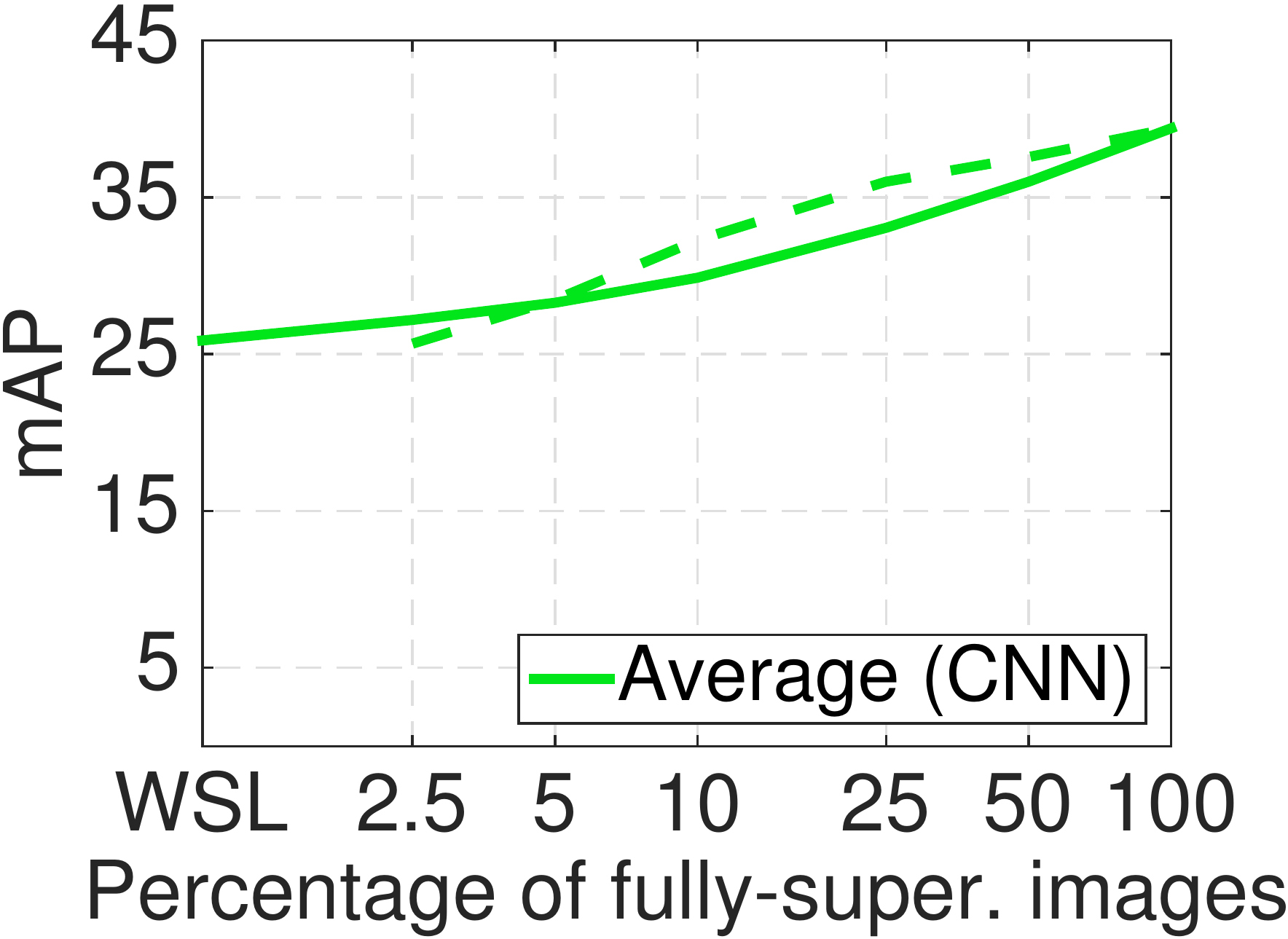} \includegraphics[width=0.19\linewidth]{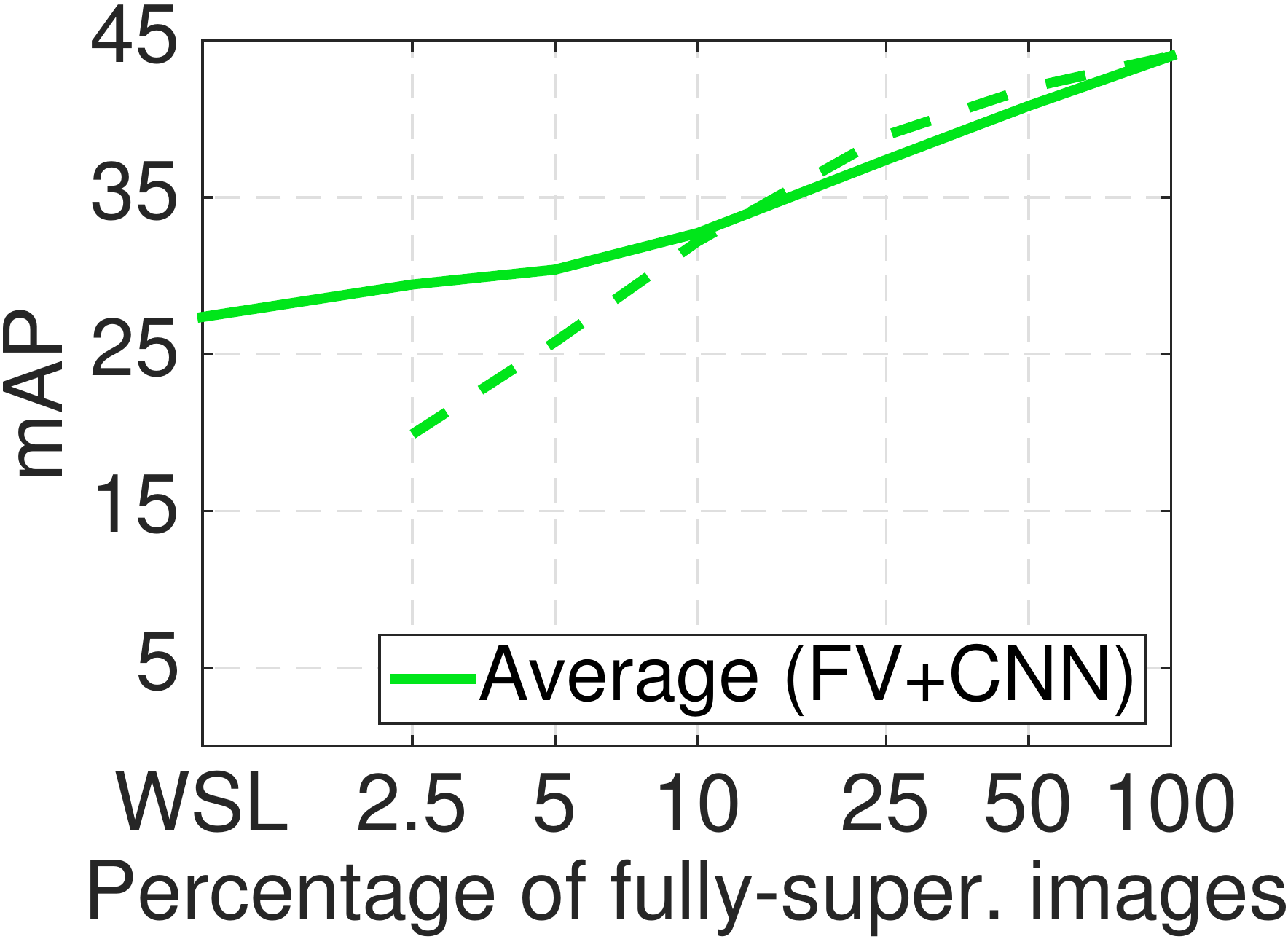}
\end{center}
\caption{Object detection results for training with mixed
supervision. Each curve shows the test set detection AP as a
function of the percentage of fully supervised positive
training images. The horizontal axes are in logarithmic scale. The first two plots show per-class curves
for selected classes using only FVs. The last
three plots show the detection AP values averaged over all
classes for the FV, CNN and FV+CNN features, respectively.
The solid curves correspond to mixed supervision. The dotted curves correspond to results obtained
by using only the fully-supervised examples.}
\label{fig:mixedsuper} 
\end{figure*}

In order to better understand the localization errors, we
categorize each of our object hypotheses in the positive
training images into one of the following five cases: (i)
correct localization (overlap $\geq$ 50\%), (ii)
hypothesis completely inside ground-truth, (iii) reversed
inclusion, (iv) none of the above, but non-zero overlap,
and (v) no overlap. For the sake of brevity, we analyze
only the WSL outputs for the FV+CNN features. In
\fig{breakdown_all_multifold} we show the frequency of
these five cases for each object category and averaged
over all classes for multi-fold MIL.  We observe that {\em hypothesis in
ground-truth} category is the second largest error mode.
For example, as expected from \fig{imagesiter:fv}, most
localization hypotheses for the class \emph{cat}, and
similarly for the class \emph{dog}, are fully contained
within a ground-truth window. Although the instances of
this mis-localization category may significantly degrade
CorLoc and AP measures, they could as well be interpreted
as correct localizations in certain applications where it
is not necessary to localize with bounding boxes fully
covering target objects. Interestingly, we observe that,
with 5.1\% on average, the ``no overlap'' case is rare.
This means that 94.9\% of our object hypotheses overlap to
some extent with a ground-truth object.  This explains the
fact that detector performance is relatively resilient to
frequent mis-localization in the sense of the CorLoc
measure.

\fig{breakdown_all_standard} presents the error
distribution corresponding to the standard MIL training.
Whereas {\em hypothesis in ground-truth} is more
frequent than {\em ground-truth in hypothesis} for
multi-fold MIL training, the situation is reversed for
standard MIL training.  This is a result of the fact that
whereas multi-fold MIL is able localize most
discriminative sub-regions of the object categories,
standard MIL tends to get stuck after the first few
iterations, resulting in too large bounding box estimates.
The effect of multi-fold training on the distribution of
different localization error types is similar when using the FV or CNN
features alone.

Finally, we note that while multi-fold MIL using $k$ folds
results in training $k$ additional classifiers per
iteration, training duration grows sublinearly with $k$
since the number of re-localizations and hard-negative
mining work does not change. In a single iteration of our
implementation using FV features, (a) all SVM
optimizations take 10.5 minutes for standard MIL and 42
minutes for 10-fold MIL, (b) re-localization on positive
images take 5 minutes in both cases and (c) hard-negative
mining takes 20 minutes in both cases. In total, standard
MIL takes 35.5 minutes per iteration and 10-fold MIL takes
67 minutes per iteration, for a single class.

\subsection{Training with mixed supervision}
\label{sec:mixedsupervision}

In our experiments so far, we have considered the WSL and
fully supervised scenarios, where each training image is
annotated with either class labels (WSL) or object
bounding boxes (fully supervised). We now consider
training using a mixture of the two paradigms, which we
refer to as {\em mixed supervision}.

One way to combine weakly supervised and fully supervised
training for object localization is to leverage an
existing dataset of fully supervised training images of
non-target classes during WSL of a new object category
detector, also referred to as transfer learning, see \eg
\cite{deselaers12ijcv,shi12bmvc}. Such an approach,
however, does not provide any fully supervised example for
the target class and does not allow hard negative mining
on the positive images, both of which are important
factors as shown in our previous analysis.

We, instead, consider a  setup where a
subset of the positive training images for each class is
fully supervised. For this purpose, we randomly sample a
subset of the positive training images and add
ground-truth box annotations for all objects in them.
These images are then excluded from the re-localization
steps in the multi-fold training procedure and instead their
ground-truth windows are used as
positive training examples. We also use the fully supervised
positive images for hard-negative mining, in addition to the negative images.

\fig{mixedsuper} presents detection AP scores as a
function of the percentage of fully supervised positive
training images. Each curve is obtained by evaluating the
performance when the ratio of
fully supervised images per class is set to values in $\{2.5,5,10,25,50,100\}\%$.
We also evaluate the baseline detection
results where only the fully supervised images are used
for training.  We repeat each experiment twice and average
the AP scores.  In each plot, the resulting mixed supervision and 
baseline curves are shown using solid and dotted lines,
respectively. The  horizontal axes are in logarithmic scale.

The leftmost panel in \fig{mixedsuper} shows the
mixed supervision evaluation results for the classes
{\em bus}, {\em horse}, {\em train}, and {\em sheep},
which we select for their similarity in performance to 
the average case for FVs (the latter is shown in the third panel). 
For these four classes, and on average, we observe a significant
performance gain using mixed supervision compared to
conventional full supervision. 

The only two classes where mixed supervision is not more
effective than fully supervised training for FVs are {\em bottle}
and {\em chair}, for which AP curves are presented in the
second panel of \fig{mixedsuper}. We note that {\em bottle} and {\em chair}
are also the classes with the lowest CorLoc scores for
multi-fold training, which explains why mixed supervised
training does not work well in these cases.

In the third panel  we observe that the fewer images are
fully supervised, the more significant the benefit of
additional weakly labeled images using FVs. Overall,
we observe that the benefit of combining
fully supervised images with weakly supervised ones is particularly significant when
the ratio of fully supervised images is up to 50\%
for FV features.

The fourth panel in \fig{mixedsuper} presents the results for
the CNN descriptors. We observe that training with mixed
supervision improves the detection mAP compared to
training with only the fully supervised examples when up
to $5\%$ of the positive training images are
fully supervised. At larger fully supervised image
percentages, training over only the fully supervised
images outperforms mixed supervision based training.
Regarding this result, we can interpret the CNN features
as the outputs from a pre-trained classifier, and
therefore, having a few training images can be sufficient
for effectively learning a detection model over the CNN
features. As a result, utilizing weakly supervised
examples during training can sometimes deteriorate the
detection performance due to the imperfect localizations
provided by the WSL methods.

Finally, the rightmost panel in \fig{mixedsuper} presents the
results for the FV+CNN combination. We observe that training with
mixed supervision is significantly beneficial
when the ratio of fully supervised examples is up to  10\%.
Above this threshold, the performance of training with 
fully supervised examples is slightly better, similar to the
CNN-only case.

Overall, the results suggest that fully supervised images
can be successfully integrated into multi-fold WSL
training in order to improve detection rates by annotating
objects only in a small number of images. This holds in
particular, when auxiliary training data, such as the
ImageNet images used for training the CNN model, is not
available. One possible direction for future work is to
give more weight to fully supervised examples
than to weakly supervised ones during
classifier training, especially in the early 
MIL iterations.

{\small\addtolength{\tabcolsep}{-3.5pt}
\begin{table*}
    \caption{Comparison of standard MIL training vs our 10-fold MIL on VOC 2010 in terms of training set localization accuracy (CorLoc).}
\centering
\begin{tabular}{|l|cccccccccccccccccccc|c|}
\cline{2-22}
\multicolumn{1}{l|}{} & aero     & bicy     & bird     & boa      & bot      & bus      & car      & cat      & cha      & cow      & dtab     & dog      & hors     & mbik     & pers     & plnt     & she      & sofa     & trai     & tv       & Av.\tabularnewline
\hline
 & \multicolumn{20}{c}{standard MIL} & \tabularnewline
\hline
FV                & {58.9}        & 45.2            & 33.7            & 24.1            & 6.7             & 66.1            & 43.3            & {{{\textbf{50.6}}}} & 16.2            & 36.0            & 25.5              & {{41.8}}      & 53.4            & 57.5            & 21.5          & 11.6                & 32.9            & 30.5            & {50.0}          & 21.6                & 36.4 \tabularnewline
\hline
CNN               & 54.8          & {{60.1}}        & 52.3            & {{40.2}}        & 26.6            & {73.9}          & 64.1            & 23.4                & 35.7            & 58.1            & 24.5              & 32.4          & {{71.3}}        & {63.8}          & 28.0          & 36.4                & 61.6            & {44.7}          & 48.1            & 55.3                & 47.8 \\
\hline
FV+CNN            & 60.2          & 53.9            & 48.5            & 34.2            & 12.6            & 71.0            & 52.6            & 44.1                & 23.3            & 37.2            & 25.5              & 45.3          & 60.2            & 61.3            & 36.0          & 15.3                & 36.6            & 34.0            & 51.0            & 31.8                & 41.7 \\
\hline
FV+CNN+Refinement & \textbf{62.7} & 56.3            & 52.8            & 39.6            & 13.5            & 71.4            & 58.7            & 47.3                & 23.9            & 44.8            & 27.7              & \textbf{54.4} & 65.9            & 66.7            & \textbf{38.0} & 19.0                & 46.8            & 34.0            & 57.8            & 38.7                & 46.0 \\
\hline
 & \multicolumn{20}{c}{multi-fold MIL} & \tabularnewline
\hline
FV                & 47.3          & 47.1            & 36.2            & 34.8            & 24.9            & 68.9            & 59.8            & 18.9                & 21.3            & 52.9            & 26.6              & 32.2          & 44.1            & 60.7            & {{33.7}}      & 17.3                & {{63.9}}        & 32.6            & 48.1            & {{{\textbf{66.6}}}} & 41.9 \tabularnewline
\hline
CNN               & 53.4          & 59.1            & {52.6}          & 39.9            & {27.1}          & 73.1            & {65.2}          & 18.6                & {40.6}          & {{68.0}}        & {33.0}            & 30.1          & 71.0            & 63.2            & 27.1          & {{{\textbf{37.8}}}} & 61.6            & 43.3            & 48.1            & 58.9                & {48.6} \\
\hline
FV+CNN            & {60.7}        & {60.1}          & {53.4}          & 38.7            & {27.8}          & {77.7}          & {67.1}          & 20.3                & {42.6}          & 64.0            & {{\textbf{39.4}}} & 38.8          & 70.6            & {65.2}          & 28.5          & 36.1                & 58.8            & {46.1}          & {55.8}          & 49.7                & {50.1} \\
\hline
FV+CNN+Refinement & {61.1}        & {\textbf{65.0}} & {\textbf{59.2}} & {\textbf{44.3}} & {\textbf{28.3}} & {\textbf{80.6}} & {\textbf{69.7}} & 31.2                & {\textbf{42.8}} & {\textbf{73.3}} & 38.3              & {50.2}        & {\textbf{74.9}} & {\textbf{70.9}} & {37.3}        & 37.1                & {\textbf{65.3}} & {\textbf{55.3}} & {\textbf{61.7}} & 58.2                & {\textbf{55.2}} \\
\hline
\end{tabular}
\label{tab:voc10_corloc}
\end{table*}
}

{\small\addtolength{\tabcolsep}{-3.5pt}
\begin{table*}
    \caption{Comparison of standard MIL training vs our 10-fold MIL on VOC 2010 in terms of test set AP measure.}
\centering
\begin{tabular}{|l|cccccccccccccccccccc|c|}
\cline{2-22}
\multicolumn{1}{l|}{} & aero          & bicy              & bird              & boa               & bot               & bus               & car               & cat                 & cha                 & cow                 & dtab                & dog           & hors              & mbik              & pers              & plnt                & she                 & sofa              & trai              & tv                  & Av.\tabularnewline
\hline
  & \multicolumn{20}{c}{standard MIL}              & \tabularnewline
\hline
FV                    & 41.9          & 30.4              & 6.9               & 5.2               & 1.6               & 38.6              & 24.8              & {{{\textbf{29.6}}}} & 1.3                 & 8.7                 & 2.3                 & 18.7          & 22.1              & 40.0              & 9.9               & 0.9                 & 9.7                 & 6.4               & 18.6              & 11.5                & 16.4 \tabularnewline
\hline
CNN                   & 35.8          & 38.6              & 21.9              & 10.1              & 8.6               & 39.0              & 33.9              & 20.5                & 8.0                 & 22.8                & 7.5                 & 17.9          & {{33.4}}          & 46.1              & 15.8              & {{{\textbf{13.6}}}} & 26.7                & 15.5              & 26.8              & 22.2                & 23.2 \\
\hline
FV+CNN                & 45.6          & 37.5              & 21.3              & 10.0              & 4.9               & 41.3              & 29.7              & 28.1                & 5.0                 & 15.5                & 7.2                 & 25.2          & 30.7              & 49.8              & 17.7              & 6.8                 & 12.2                & 10.9              & 28.5              & 9.8                 & 21.9 \\
\hline
FV+CNN+Refinement     & \textbf{47.3} & 37.3              & 24.1              & 11.0              & 5.6               & 41.9              & 31.9              & 27.9                & 5.1                 & 15.2                & 7.7                 & \textbf{29.9} & 32.0              & 52.2              & \textbf{20.7}     & 8.4                 & 15.9                & 12.7              & 30.8              & 13.0                & 23.5 \\
\hline
  & \multicolumn{20}{c}{multi-fold MIL}            & \tabularnewline
\hline
FV                    & 27.9          & 23.2              & 8.1               & {{11.8}}          & {{9.6}}           & 35.7              & 31.3              & 10.7                & 3.6                 & 14.9                & 6.0                 & 12.8          & 18.6              & 41.8              & {{16.3}}          & 3.0                 & {{{\textbf{27.6}}}} & 10.3              & 22.4              & {{{\textbf{34.6}}}} & 18.5 \tabularnewline
\hline
CNN                   & 34.7          & 39.1              & 21.9              & 10.5              & 8.8               & 37.7              & 34.4              & 18.1                & 10.1                & {{{\textbf{26.4}}}} & 11.2                & 16.5          & 33.0              & 44.7              & 15.6              & 13.2                & 26.2                & 15.6              & 24.8              & 24.8                & 23.4 \\
\hline
FV+CNN                & {{42.2}}      & {{41.5}}          & {{22.5}}          & 11.3              & 8.6               & {{41.7}}          & {{36.1}}          & 19.4                & {{{\textbf{13.3}}}} & 24.3                & {{{\textbf{14.5}}}} & {{21.3}}      & 32.7              & {{48.3}}          & 15.2              & 11.3                & 25.0                & {{18.0}}          & {{27.9}}          & 18.4                & {{24.7}} \\
\hline
FV+CNN+Refinement     & {{44.6}}      & {{\textbf{42.3}}} & {{\textbf{25.5}}} & {{\textbf{14.1}}} & {{\textbf{11.0}}} & {{\textbf{44.1}}} & {{\textbf{36.3}}} & {23.2}              & {12.2}              & {26.1}              & {14.0}              & {{29.2}}      & {{\textbf{36.0}}} & {{\textbf{54.3}}} & {{\textbf{20.7}}} & {12.4}              & {26.5}              & {{\textbf{20.3}}} & {{\textbf{31.2}}} & {23.7}              & {{\textbf{27.4}}} \\
\hline
\end{tabular}
\label{tab:voc10_ap}
\end{table*}
}

\subsection{VOC 2010 evaluation}
\label{sec:voc10}

We now present an evaluation on the VOC 2010 dataset in
order to verify the effectiveness of multi-fold training
and the window refinement method on a second dataset.  We
are the first to present weakly supervised results on this
dataset, and can therefore not compare to other weakly
supervised methods.  We show the resulting CorLoc values
in \tab{voc10_corloc} and detection AP results in
\tab{voc10_ap}. Overall, our results on VOC 2010 are
similar to those on the 2007 dataset in the sense that
multi-fold MIL significantly improves the WSL performance
compared to standard MIL training, especially when
high-dimensional FV descriptors are included.  Using
multi-fold MIL over the combined FV and CNN features
results in 24.7\% mAP, which is significantly better than 
21.9\% mAP by standard MIL. The window refinement method
further improves multi-fold MIL performance from 24.7\% to
27.4\% mAP.

If we train the object detectors in a fully supervised
manner, we obtain 33.6\% mAP using the FV features, and
37.7\% mAP using the CNN features. This verifies that we
have an effective object representation outperforming
DPMs~\cite{felzdet_ver5} (29.6\% mAP). On this dataset,
the highest fully supervised detection result without
using auxiliary data is 39.7\% mAP~\cite{wang13iccvRegionlets}.
We note that whereas the
CNN model we use is trained on the ImageNet images only,
Girshick \etal~\cite{girshick14cvpr} utilize a CNN model
{\em fine-tuned} on the VOC ground-truth boxes, which
leads to a better fully-supervised detection performance
of 53.7\% mAP. We plan to explore weakly supervised CNN
fine-tuning in future work.

\section{Conclusions}
\label{sec:conclusion}

In this article, we have introduced a multi-fold multiple instance learning
approach for weakly supervised object detection, which
avoids the degenerate localization performance observed
without it.  Second, we have presented a contrastive background
descriptor, which encourages the detection model to learn
the differences between the objects and their context.
Third, we have designed a window refinement method, which
improves the localization accuracy by using an
edge-driven objectness prior.

We have evaluated our approach and compared it to
state-of-the-art methods using the VOC 2007 dataset.  Our
results show that multi-fold MIL effectively handles
high-dimensional descriptors, which allows us to obtain
state-of-the-art results by jointly using FV and
CNN features. On the VOC 2010 dataset we observe
similar improvements by using our multi-fold MIL
method.

A detailed analysis of our results shows that, in terms of
test set detection performance, multi-fold MIL attains
68\% of the MIL performance upper-bound, which we measure 
by selecting one correct training example from each positive 
image, for the combined FV and CNN features.

\smallskip

\noindent {\bf Acknowledgements.} This work was supported by the
European integrated project AXES and the ERC advanced grant ALLEGRO.

\bibliographystyle{ieee}
\bibliography{bibabbr,jjv.copy,rgc}

\begin{thebibliography}{10}\itemsep=-1pt

\bibitem{alexe10cvpr}
B.~Alexe, T.~Deselaers, and V.~Ferrari.
\newblock What is an object?
\newblock In {\em IEEE Conference on Computer Vision and Pattern Recognition},
  2010.

\bibitem{alexe12pami}
B.~Alexe, T.~Deselaers, and V.~Ferrari.
\newblock Measuring the objectness of image windows.
\newblock {\em IEEE Transactions on Pattern Analysis and Machine Intelligence},
  34(11):2189--2202, 2012.

\bibitem{alted10cse}
F.~Alted.
\newblock Why modern {CPU}s are starving and what can be done about it.
\newblock {\em Computing in Science \& Engineering}, 12(2):68--71, 2010.

\bibitem{bagon10cvpr}
S.~Bagon, O.~Brostovski, M.~Galun, and M.~Irani.
\newblock Detecting and sketching the common.
\newblock In {\em IEEE Conference on Computer Vision and Pattern Recognition},
  2010.

\bibitem{bengio09icml}
Y.~Bengio, J.~Louradour, R.~Collobert, and J.~Weston.
\newblock Curriculum learning.
\newblock In {\em International Conference on Machine Learning}, 2009.

\bibitem{berg04cvpr}
T.~Berg, A.~Berg, J.~Edwards, M.~Maire, R.~White, Y.~Teh, E.~Learned-Miller,
  and D.~Forsyth.
\newblock Names and faces in the news.
\newblock In {\em IEEE Conference on Computer Vision and Pattern Recognition},
  2004.

\bibitem{bilen14ijcv}
H.~Bilen, V.~Namboodiri, and L.~{Van Gool}.
\newblock Object and action classification with latent window parameters.
\newblock {\em International Journal on Computer Vision}, 106(3):237--251,
  2014.

\bibitem{bilen14bmvc}
H.~Bilen, M.~Pedersoli, and T.~Tuytelaars.
\newblock Weakly supervised object detection with posterior regularization.
\newblock In {\em British Machine Vision Conference}, 2014.

\bibitem{blaschko10nips}
M.~Blaschko, A.~Vedaldi, and A.~Zisserman.
\newblock Simultaneous object detection and ranking with weak supervision.
\newblock In {\em Advances in Neural Information Processing Systems}, 2010.

\bibitem{blei03jmlr}
D.~Blei, A.~Ng, and M.~Jordan.
\newblock Latent {D}irichlet allocation.
\newblock {\em Journal of Machine Learning Research}, 3:993--1022, 2003.

\bibitem{cho15cvpr}
M.~Cho, S.~Kwak, C.~Schmid, and J.~Ponce.
\newblock Unsupervised {Object} {Discovery} and {Localization} in the {Wild}:
  {Part}-based {Matching} with {Bottom}-up {Region} {Proposals}.
\newblock In {\em {CVPR}}, 2015.

\bibitem{chum07cvpr}
O.~Chum and A.~Zisserman.
\newblock An exemplar model for learning object classes.
\newblock In {\em IEEE Conference on Computer Vision and Pattern Recognition},
  2007.

\bibitem{cinbis13iccv}
R.~Cinbis, J.~Verbeek, and C.~Schmid.
\newblock Segmentation driven object detection with {F}isher vectors.
\newblock In {\em International Conference on Computer Vision}, 2013.

\bibitem{cinbis14cvpr}
R.~Cinbis, J.~Verbeek, and C.~Schmid.
\newblock Multi-fold {MIL} training for weakly supervised object localization.
\newblock In {\em IEEE Conference on Computer Vision and Pattern Recognition},
  2014.

\bibitem{crandall06eccv}
D.~Crandall and D.~Huttenlocher.
\newblock Weakly supervised learning of part-based spatial models for visual
  object recognition.
\newblock In {\em European Conference on Computer Vision}, 2006.

\bibitem{dance04eccv}
G.~Csurka, C.~Dance, L.~Fan, J.~Willamowski, and C.~Bray.
\newblock Visual categorization with bags of keypoints.
\newblock In {\em ECCV Int. Workshop on Stat. Learning in Computer Vision},
  2004.

\bibitem{deselaers12ijcv}
T.~Deselaers, B.~Alexe, and V.~Ferrari.
\newblock Weakly supervised localization and learning with generic knowledge.
\newblock {\em International Journal on Computer Vision}, 100(3):257--293,
  2012.

\bibitem{dietterich97ai}
T.~Dietterich, R.~Lathrop, and T.~Lozano-P{\'e}rez.
\newblock Solving the multiple instance problem with axis-parallel rectangles.
\newblock {\em Artificial Intelligence}, 89(1-2):31--71, 1997.

\bibitem{everingham09ivc}
M.~Everingham, J.~Sivic, and A.~Zisserman.
\newblock Taking the bite out of automatic naming of characters in {TV} video.
\newblock {\em Image and Vision Computing}, 27(5):545--559, 2009.

\bibitem{everingham10ijcv}
M.~Everingham, L.~{van Gool}, C.~Williams, J.~Winn, and A.~Zisserman.
\newblock The {PASCAL} visual object classes ({VOC}) challenge.
\newblock {\em International Journal on Computer Vision}, 88(2):303--338, 2010.

\bibitem{felzenszalb10pami}
P.~Felzenszwalb, R.~Grishick, D.~{McAllester}, and D.~Ramanan.
\newblock Object detection with discriminatively trained part based models.
\newblock {\em IEEE Transactions on Pattern Analysis and Machine Intelligence},
  32(9), 2010.

\bibitem{girshick14cvpr}
R.~Girshick, J.~Donahue, T.~Darrell, and J.~Malik.
\newblock Rich feature hierarchies for accurate object detection and semantic
  segmentation.
\newblock In {\em IEEE Conference on Computer Vision and Pattern Recognition},
  2013.

\bibitem{felzdet_ver5}
R.~Girshick, P.~Felzenszwalb, and D.~McAllester.
\newblock Discriminatively trained deformable part models, release 5.
\newblock http://people.cs.uchicago.edu/~rbg/latent-release5, 2012.

\bibitem{gu12eccv}
C.~Gu, P.~Arbel\'aez, Y.~Lin, K.~Yu, and Malik.
\newblock Multi-component models for object detection.
\newblock In {\em European Conference on Computer Vision}, 2012.

\bibitem{hofmann01ml}
T.~Hofmann.
\newblock Unsupervised learning by probabilistic latent semantic analysis.
\newblock {\em Machine Learning}, 42(1/2):177--196, 2001.

\bibitem{jegou11pami2}
H.~J{\'e}gou, M.~Douze, and C.~Schmid.
\newblock Product quantization for nearest neighbor search.
\newblock {\em IEEE Transactions on Pattern Analysis and Machine Intelligence},
  33(1):117--128, 2011.

\bibitem{jia14caffe}
Y.~Jia, E.~Shelhamer, J.~Donahue, S.~Karayev, J.~Long, R.~Girshick,
  S.~Guadarrama, and T.~Darrell.
\newblock Caffe: Convolutional architecture for fast feature embedding.
\newblock {\em arXiv preprint arXiv:1408.5093}, 2014.

\bibitem{kim09nips}
G.~Kim and A.~Torralba.
\newblock Unsupervised detection of regions of interest using iterative link
  analysis.
\newblock In {\em Advances in Neural Information Processing Systems}, pages
  4--2, 2009.

\bibitem{krizhevsky12nips}
A.~Krizhevsky, I.~Sutskever, and G.~Hinton.
\newblock Imagenet classification with deep convolutional neural networks.
\newblock In {\em Advances in Neural Information Processing Systems}, pages
  1106--1114, 2012.

\bibitem{kumar10nips}
M.~P. Kumar, B.~Packer, and D.~Koller.
\newblock Self-paced learning for latent variable models.
\newblock In {\em Advances in Neural Information Processing Systems}, 2010.

\bibitem{lampert09pami}
C.~Lampert, M.~Blaschko, and T.~Hofmann.
\newblock Efficient subwindow search: a branch and bound framework for object
  localization.
\newblock {\em IEEE Transactions on Pattern Analysis and Machine Intelligence},
  31(12):2129--2142, 2009.

\bibitem{lazebnik06cvpr}
S.~Lazebnik, C.~Schmid, and J.~Ponce.
\newblock Beyond bags of features: spatial pyramid matching for recognizing
  natural scene categories.
\newblock In {\em IEEE Conference on Computer Vision and Pattern Recognition},
  2006.

\bibitem{nesterov05mathprog}
Y.~Nesterov.
\newblock Smooth minimization of non-smooth functions.
\newblock {\em Mathematical programming}, 103(1):127--152, 2005.

\bibitem{nguyen09iccv}
M.~Nguyen, L.~Torresani, F.~de~la Torre, and C.~Rother.
\newblock Weakly supervised discriminative localization and classification: a
  joint learning process.
\newblock In {\em International Conference on Computer Vision}, 2009.

\bibitem{pandey11iccv}
M.~Pandey and S.~Lazebnik.
\newblock Scene recognition and weakly supervised object localization with
  deformable part-based models.
\newblock In {\em International Conference on Computer Vision}, 2011.

\bibitem{parkhi11iccv}
O.~Parkhi, A.~Vedaldi, C.~Jawahar, and A.~Zisserman.
\newblock The truth about cats and dogs.
\newblock In {\em International Conference on Computer Vision}, 2011.

\bibitem{prest12cvpr}
A.~Prest, C.~Leistner, J.~Civera, C.~Schmid, and V.~Ferrari.
\newblock Learning object class detectors from weakly annotated video.
\newblock In {\em IEEE Conference on Computer Vision and Pattern Recognition},
  2012.

\bibitem{russakovsky12eccv}
O.~Russakovsky, Y.~Lin, K.~Yu, and L.~Fei-Fei.
\newblock Object-centric spatial pooling for image classification.
\newblock In {\em European Conference on Computer Vision}, 2012.

\bibitem{sanchez13ijcv}
J.~S\'{a}nchez, F.~Perronnin, T.~Mensink, and J.~Verbeek.
\newblock Image classification with the {F}isher vector: Theory and practice.
\newblock {\em International Journal on Computer Vision}, 105(3):222--245,
  2013.

\bibitem{shi13iccv}
Z.~Shi, T.~Hospedales, and T.~Xiang.
\newblock Bayesian joint topic modelling for weakly supervised object
  localisation.
\newblock In {\em International Conference on Computer Vision}, 2013.

\bibitem{shi12bmvc}
Z.~Shi, P.~Siva, T.~Xiang, and Q.~Mary.
\newblock Transfer learning by ranking for weakly supervised object annotation.
\newblock In {\em {BMVC}}, 2012.

\bibitem{singh12eccv}
S.~Singh, A.~Gupta, and A.~Efros.
\newblock Unsupervised discovery of mid-level discriminative patches.
\newblock In {\em European Conference on Computer Vision}, 2012.

\bibitem{siva12eccv}
P.~Siva, C.~Russell, and T.~Xiang.
\newblock In defence of negative mining for annotating weakly labelled data.
\newblock In {\em European Conference on Computer Vision}, 2012.

\bibitem{siva13cvpr}
P.~Siva, C.~Russell, T.~Xiang, and L.~Agapito.
\newblock Looking beyond the image: Unsupervised learning for object saliency
  and detection.
\newblock In {\em IEEE Conference on Computer Vision and Pattern Recognition},
  2013.

\bibitem{siva11iccv}
P.~Siva and T.~Xiang.
\newblock Weakly supervised object detector learning with model drift
  detection.
\newblock In {\em International Conference on Computer Vision}, 2011.

\bibitem{song14icml}
H.~Song, R.~Girshick, S.~Jegelka, J.~Mairal, Z.~Harchaoui, and T.~Darrell.
\newblock On learning to localize objects with minimal supervision.
\newblock In {\em International Conference on Machine Learning}, 2014.

\bibitem{song14nips}
H.~Song, Y.~Lee, S.~Jegelka, and T.~Darrell.
\newblock Weakly-supervised discovery of visual pattern configurations.
\newblock In {\em Advances in Neural Information Processing Systems}, 2014.

\bibitem{song2011cvpr}
Z.~Song, Q.~Chen, Z.~Huang, Y.~Hua, and S.~Yan.
\newblock Contextualizing object detection and classification.
\newblock In {\em IEEE Conference on Computer Vision and Pattern Recognition},
  2011.

\bibitem{uijlings13ijcv}
J.~Uijlings, K.~van~de Sande, T.~Gevers, and A.~Smeulders.
\newblock Selective search for object recognition.
\newblock {\em International Journal on Computer Vision}, 104(2):154--171,
  2013.

\bibitem{sande14cvpr}
K.~van~de Sande, C.~Snoek, and A.~Smeulders.
\newblock Fisher and {VLAD} with {FLAIR}.
\newblock In {\em IEEE Conference on Computer Vision and Pattern Recognition},
  2014.

\bibitem{verbeek07cvpr}
J.~Verbeek and B.~Triggs.
\newblock Region classification with {M}arkov field aspect models.
\newblock In {\em IEEE Conference on Computer Vision and Pattern Recognition},
  2007.

\bibitem{vijayanarasimhan11cvpr}
S.~Vijayanarasimhan and K.~Grauman.
\newblock Large-scale live active learning: Training object detectors with
  crawled data and crowds.
\newblock In {\em IEEE Conference on Computer Vision and Pattern Recognition},
  2011.

\bibitem{wang14eccv}
C.~Wang, W.~Ren, K.~Huang, and T.~Tan.
\newblock Weakly supervised object localization with latent category learning.
\newblock In {\em European Conference on Computer Vision}, 2014.

\bibitem{wang10cvpr}
J.~Wang, J.~Yang, K.~Yu, F.~Lv, T.~Huang, and Y.~Gong.
\newblock Locality-constrained linear coding for image classification.
\newblock In {\em IEEE Conference on Computer Vision and Pattern Recognition},
  2010.

\bibitem{wang13iccvRegionlets}
X.~Wang, M.~Yang, S.~Zhu, and Y.~Lin.
\newblock Regionlets for generic object detection.
\newblock In {\em International Conference on Computer Vision}, 2013.

\bibitem{zitnick14eccv}
C.~Zitnick and P.~Doll\'{a}r.
\newblock Edge boxes: Locating object proposals from edges.
\newblock In {\em European Conference on Computer Vision}, 2014.

\end{thebibliography}

\vspace{-5.5mm}
\begin{IEEEbiography}[{\includegraphics[width=1in,height=1.25in,clip,keepaspectratio]{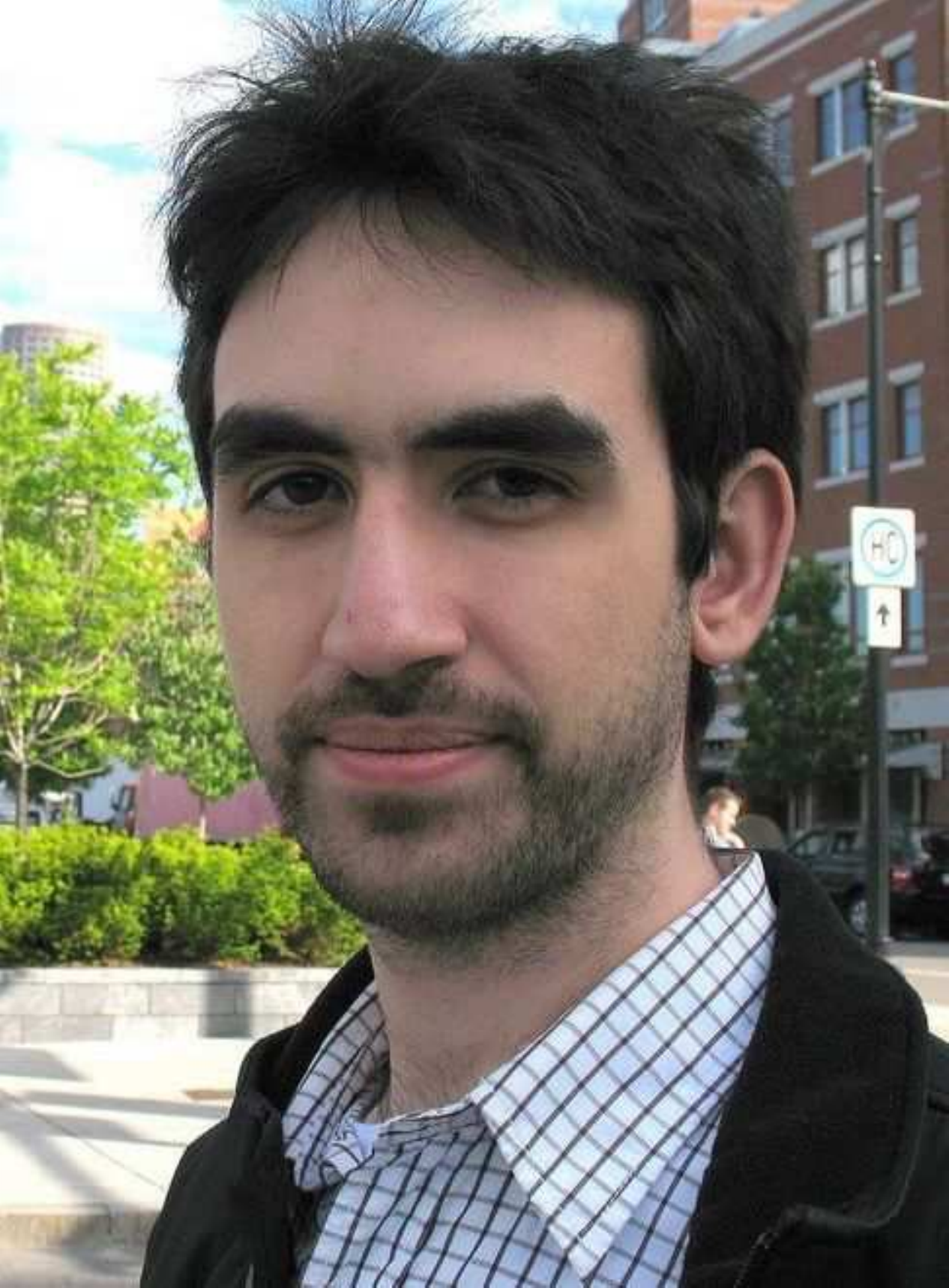}}]{Ramazan Gokberk Cinbis}
graduated from Bilkent University, Turkey, in 2008, and
received an M.A. degree from Boston University, USA, in 2010.
He was
a doctoral student in the LEAR team, at INRIA Grenoble,
from 2010 until 2014, and received a PhD degree in
computer science from Universit\'{e} de Grenoble, France,
in 2014. 
He is currently an assistant professor at the Department of Computer Engineering, Bilkent University, Ankara, Turkey. 
His research interests include computer vision and machine learning.  
\end{IEEEbiography}

\vspace{-3.5mm}

\begin{IEEEbiography}[{\includegraphics[width=1in,height=1.25in,clip,keepaspectratio]{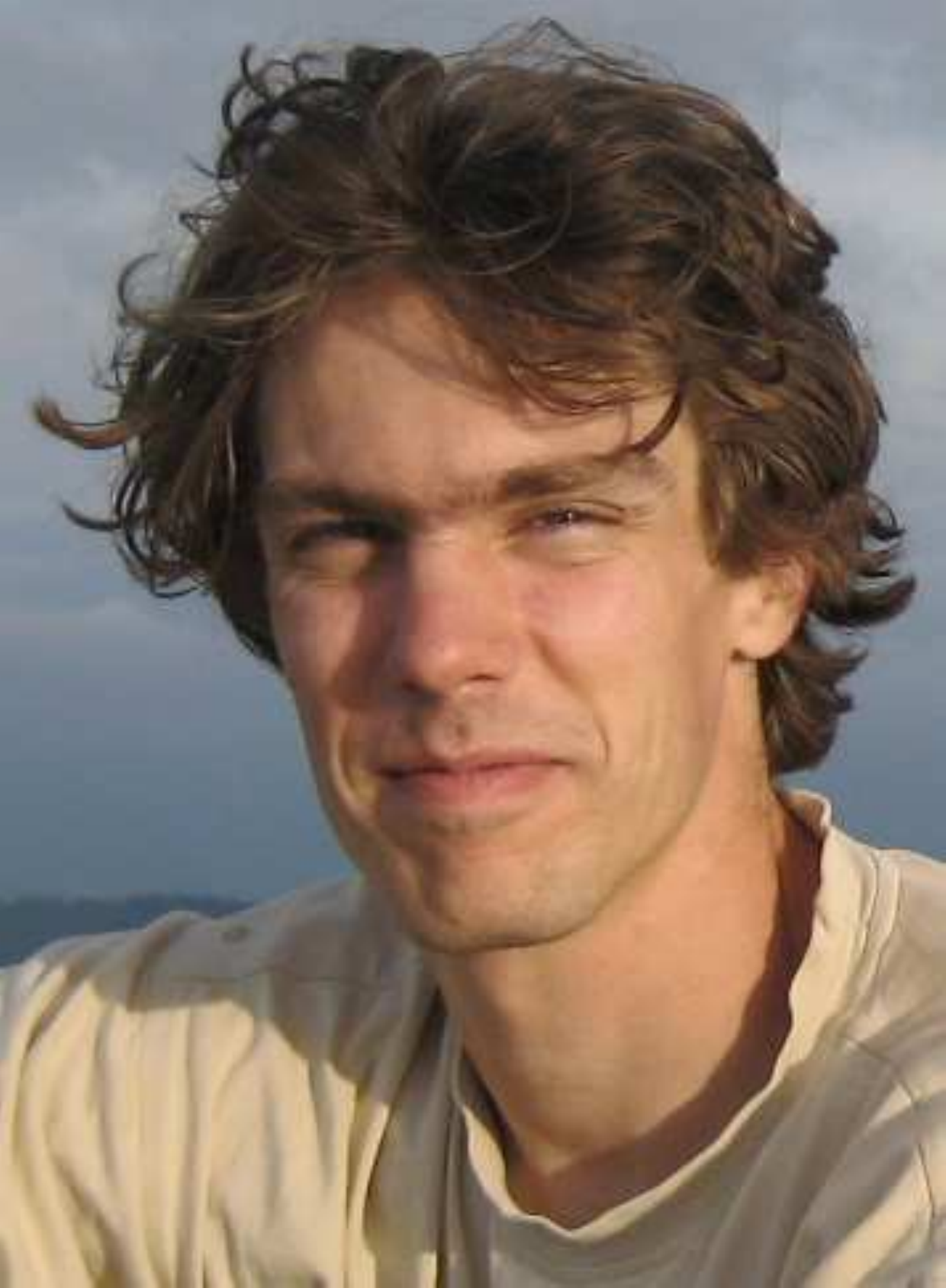}}]{Jakob Verbeek}
received a PhD degree in computer science in 2004 from the University of Amsterdam, The Netherlands. 
After being a postdoctoral researcher at the University of Amsterdam and at INRIA Rh\^one-Alpes, he has
been a full-time researcher  at INRIA, Grenoble, France, since 2007. 
His research interests include machine learning and computer vision, with special interest in applications of
statistical models in computer vision. 
\end{IEEEbiography}

\vspace{-3.5mm}

\begin{IEEEbiography}[{\includegraphics[width=1in,height=1.25in,clip,keepaspectratio]{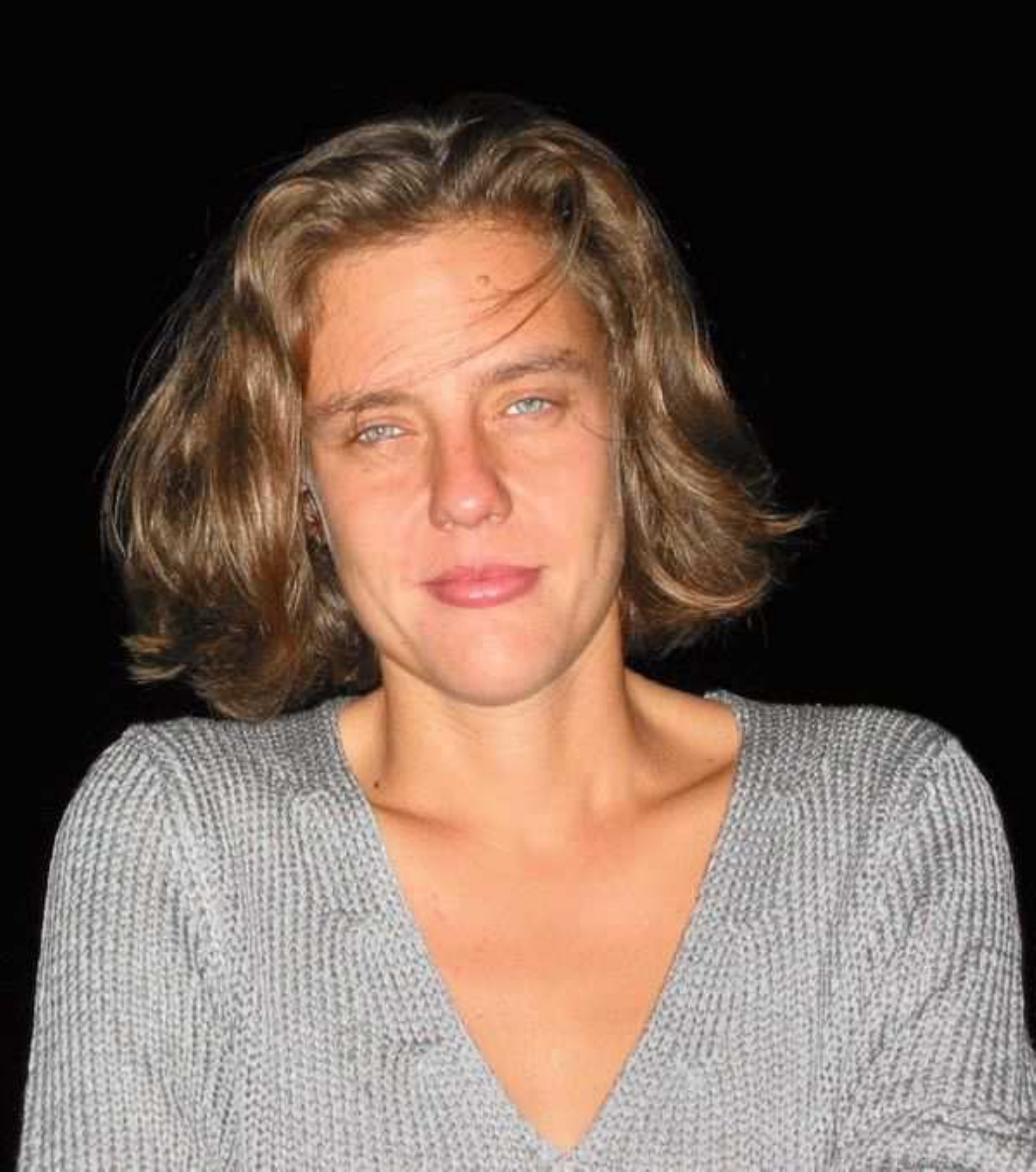}}]{Cordelia Schmid}
holds a M.S. degree in computer science
from the University of Karlsruhe and a doctorate from the
Institut National Polytechnique de Grenoble. She is a
research director at INRIA Grenoble where she directs the
LEAR team. In 2006 and 2014, she was awarded the Longuet-Higgins
prize for fundamental contributions in computer vision
that have withstood the test of time. In 2012, she
obtained an ERC advanced grant. She is a fellow of IEEE.
\end{IEEEbiography}

\end{document}